\documentclass[letterpaper, 10 pt, conference]{ieeeconf}  

\IEEEoverridecommandlockouts                              

\usepackage{graphics} 
\usepackage{epsfig} 
\usepackage{mathptmx} 
\usepackage{times} 
\usepackage{amsmath} 
\usepackage{amssymb}  
\usepackage{url}
\usepackage{subcaption}
\usepackage{xspace}
\usepackage{tabularx}
\usepackage{graphicx}
\usepackage{multirow}
\usepackage{xparse}
\usepackage{nicefrac}
\usepackage{microtype}
\usepackage{amsfonts}
\usepackage{enumerate}
\usepackage{setspace}

\let\labelindent\relax

\usepackage{enumitem}
\usepackage{wrapfig}
\usepackage{booktabs} %
\usepackage{tikz}

\usepackage{amsbsy}
\usepackage{array} 
\usepackage{booktabs}
\usepackage{array}
\usepackage{caption}
\usepackage{algorithm}
\usepackage{algpseudocode}
\usepackage{makecell}

\usepackage{listings}
\usepackage{xcolor}
\usepackage{tcolorbox}
\tcbuselibrary{listingsutf8}

\lstdefinestyle{pythoncode}{
    language=Python,
    basicstyle=\ttfamily\scriptsize, 
    keywordstyle=\color{blue},
    commentstyle=\color{gray},
    stringstyle=\color{orange},
    backgroundcolor=\color{gray!10}, 
    frame=single, 
    breaklines=true, 
    captionpos=b, 
    numbers=left, 
    numberstyle=\tiny\color{gray}, 
    stepnumber=1, 
    numbersep=6pt, 
    xleftmargin=12pt, 
    showspaces=false, 
    showstringspaces=false, 
    showtabs=false, 
    tabsize=4, 
    keepspaces=true, 
    columns=flexible 
}

\lstdefinestyle{plaintext}{
    basicstyle=\ttfamily\scriptsize, 
    backgroundcolor=\color{gray!10}, 
    frame=single, 
    breaklines=true, 
    breakatwhitespace=false, 
    prebreak=, 
    postbreak=, 
    captionpos=b, 
    numbers=none, 
    showspaces=false, 
    showstringspaces=false, 
    showtabs=false, 
    tabsize=4, 
    xleftmargin=0pt, 
    xrightmargin=0pt, 
    columns=fullflexible, 
    keepspaces=true, 
    breakindent=0pt, 
    resetmargins=true 
}

\newcommand{\methodname}{ZeroMimic\xspace}

\newcommand{\overallrealsuccessrate}{71.0\%\xspace}
\newcommand{\overallfrankasuccessrate}{71.9\%\xspace}
\newcommand{\overallwidowxsuccessrate}{65.0\%\xspace}
\newcommand{\overallsimsuccessrate}{73.8\%\xspace}
\newcommand{\numskills}{9\xspace}
\newcommand{\numobjects}{18\xspace}
\newcommand{\numrobots}{2\xspace}
\newcommand{\numrealinstances}{30\xspace}
\newcommand{\numtotalinstances}{34\xspace}
\newcommand{\numrealkitchenscenes}{6\xspace}

\makeatletter
\let\NAT@parse\undefined
\makeatother
\usepackage[numbers,sort&compress]{natbib}
\usepackage{hyperref}

\title{\LARGE \bf
\methodname:
Distilling Robotic Manipulation Skills from Web Videos \\
}

\author{Junyao Shi$^{*}$, Zhuolun Zhao$^{*}$, Tianyou Wang, Ian Pedroza$^{\dag}$, Amy Luo$^{\dag}$, Jie Wang, Jason Ma, Dinesh Jayaraman
\\
University of Pennsylvania
\\
\href{https://zeromimic.github.io}{\textbf{zeromimic.github.io}}
\\
\thanks{Email correspondence: junys@seas.upenn.edu}%
\thanks{$^{*}$ \jsadd{and $^{\dagger}$ denote} equal contribution.}%
\\[-4.0ex]
}

\usepackage[addedmarkup=uline, defaultcolor=magenta, todonotes={textsize=scriptsize, textwidth=2cm}, authormarkuptext=name, commandnameprefix=always, xcolor, final]{changes}
\definechangesauthor[name={JD}, color=orange]{jd}
\definechangesauthor[name={JS}, color=blue]{js}
\definechangesauthor[name={A}, color=blue]{ax}

\newcommand{\jdadd}[1]{\chadded[id=jd]{#1}}

\newcommand{\jsadd}[1]{\chadded[id=js]{#1}}

\newcommand{\axadd}[1]{\chadded[id=ax]{#1}}
\newcommand{\axdrop}[1]{\chdeleted[id=ax]{#1}}

\begin{document}

\maketitle
\thispagestyle{empty}
\pagestyle{empty}

\raggedbottom
\begin{figure*}[!bt]
\centering 
\includegraphics[width=0.85\textwidth]{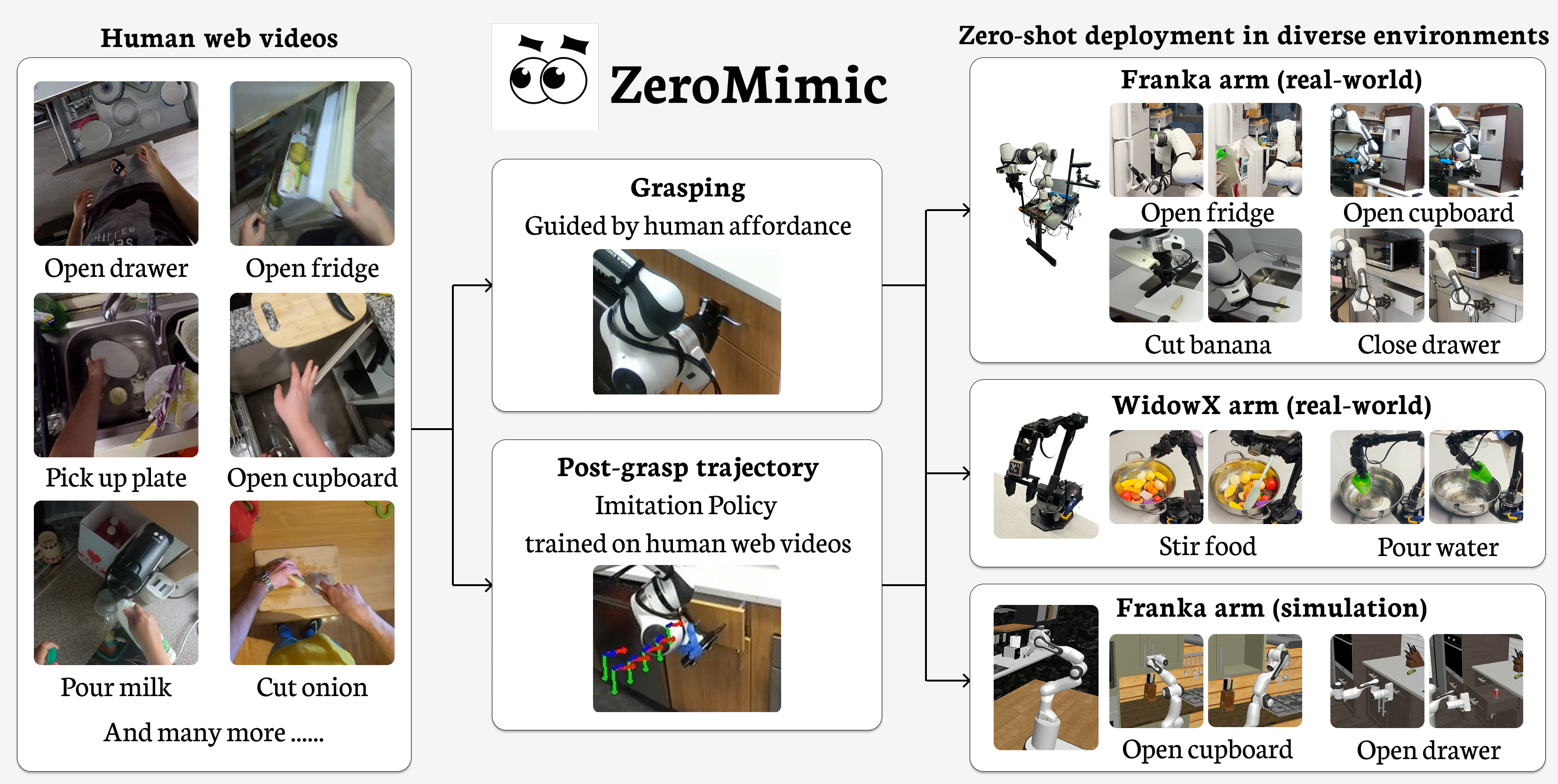}
\caption{\textbf{\methodname} distills robotic manipulation skills from egocentric web videos for zero-shot deployment \jsadd{across diverse real-world and simulated environments, a variety of objects, and different robot embodiments}.
}
\vspace{-0.4cm}
\label{fig:overview}
\end{figure*}
\vspace{-3mm}
\begin{abstract}
Many recent advances in robotic manipulation have come through imitation learning, yet these rely largely on mimicking a particularly hard-to-acquire form of demonstrations: those collected on the same robot in the same room with the same objects as the trained policy must handle at test time. In contrast, large pre-recorded human video datasets demonstrating manipulation skills in-the-wild already exist, which contain valuable information for robots.
Is it possible to distill a repository of useful robotic skill policies out of such data without any additional requirements on robot-specific demonstrations or exploration? 
We present the first such system \methodname, that generates immediately deployable image goal-conditioned skill policies for several common categories of manipulation tasks (opening, closing, pouring, pick\&place, cutting, and stirring) each capable of acting upon diverse objects and across diverse unseen task setups. \methodname is carefully designed to exploit recent advances in semantic and geometric visual understanding of human videos, together with modern grasp affordance detectors and imitation policy classes. After training \methodname on the popular EpicKitchens dataset of ego-centric human videos, we evaluate its out-of-the-box performance in varied \axadd{real-world and simulated} kitchen settings \axadd{with two different robot embodiments}, 
demonstrating its impressive abilities to handle these varied tasks. To enable plug-and-play reuse of \methodname policies on other task setups and robots, \axadd{we release software and policy checkpoints of our skill policies}.
\end{abstract}


\section{Introduction}

It is clear that animals and humans are able to observe third-person experiences to acquire functional sensorimotor skills, often ``zero-shot'' with limited or no need for additional practice. For example, one can learn to cook pasta, use a wood lathe, plant a garden, or tie a necktie, with reasonable proficiency by watching how-to video demonstrations on the web. While ``imitation learning'' has also been instrumental in many recent successes for \textit{robotic} manipulation \cite{aloha, diffusionpolicy, mobilealoha, alohaunleashed}, these robots largely rely on a much stronger kind of demonstration --- gathered by manually operating the very same robot in the same small set of scenarios (scenes, viewpoints, objects, lighting, background textures, and distractors) to perform the task of interest. 
This is an immediate stumbling block on the road to developing general-purpose robots: gathering robot- and scenario-specific demonstrations scales poorly.

Learning robot skills from in-the-wild human videos offers the enticing prospect that data would no longer be a bottleneck: videos of humans demonstrating varied manipulation tasks in diverse scenarios are already available on the web, it is easy to gather many more if needed, and further, the same videos could be re-used for many robots. However, there are serious challenges. Robots differ from humans in embodiments, action spaces, and hardware capabilities. Individual web videos often do not conveniently present all the details of how to perform a task (e.g. occlusions, out-of-frame objects and actions, or shaky moving cameras).  Finally, the distribution of in-the-wild videos spans very large variations that may be hard to handle.

We present an approach, \methodname, that systematically overcomes these challenges and distills in-the-wild egocentric videos from EpicKitchens~\cite{epickitchens} into a repository of off-the-shelf deployable image goal-conditioned robotic manipulation skill policies that transfer across scenarios. Briefly, we abstract the action spaces of humans and standard robot arms with two-fingered grippers to permit coarse action transfer, we exploit video activity understanding and pre-existing visuomotor robot primitives such as grasping to transfer the finer details of control, we exploit modern structure-from-motion systems to maintain 3D maps of noisy and shaky in-the-wild egocentric human videos, and demonstrate that large policy classes can digest the diversity of web video to learn useful behaviors. The resulting system empirically demonstrates zero-shot robotic manipulation capabilities to perform a wide range of skills with diverse objects.
In summary, our contributions are:
\begin{enumerate}
    \item We develop \methodname, a system that distills robotic manipulation skills from web videos that can be deployed zero-shot in diverse everyday environments.
    \item We evaluate \methodname on \numskills different skills and show that \methodname achieves \overallrealsuccessrate 
    out-of-the-box success rate \axadd{in the real world, \overallsimsuccessrate success rate in simulation}, can generalize to new objects unseen in our curated web video\axadd{, and can be deployed across different robot embodiments}. 
    \item Our ablation studies reveal important lessons of what is important in learning and executing robotic skills purely from in-the-wild human videos. 
\end{enumerate}

\section{Related Work}
\label{sec:related-works}

\begin{figure}
\centering 
\includegraphics[width=0.45\textwidth]{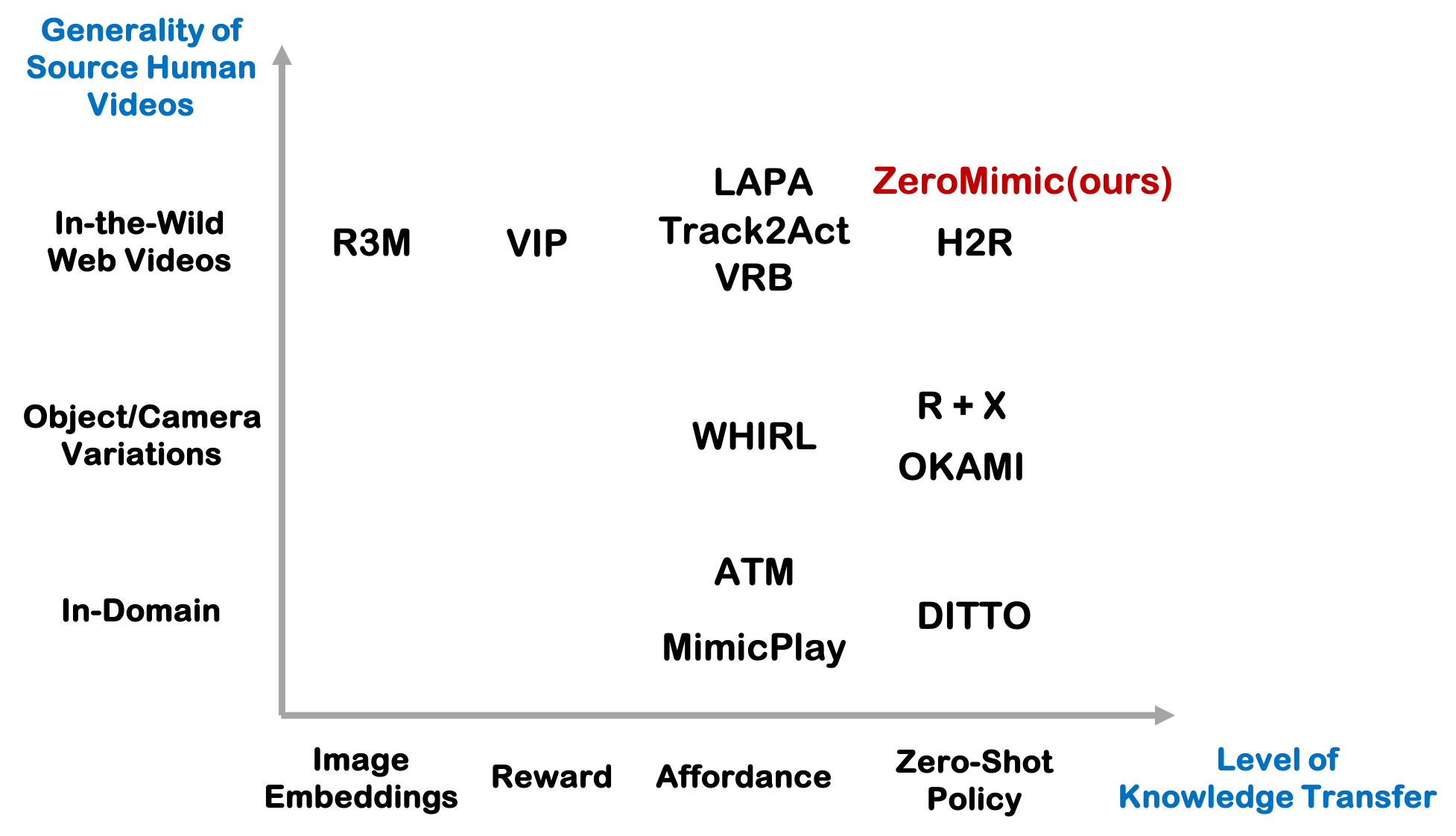}
\caption{Representative related work organized by \textbf{Generality of Source Human Videos} and \textbf{Level of Knowledge Transfer}. \methodname learns diverse zero-shot policies from in-the-wild web videos.}
\vspace{-0.5cm}
\label{fig:related_work_map}
\end{figure}

Popular recent approaches~\cite{aloha, diffusionpolicy, mobilealoha} for enabling robot manipulation often rely on costly high-quality in-domain robot demonstrations. Therefore, recent works in robot learning have increasingly focused on leveraging unstructured or out-of-domain data. Some works have demonstrated the zero-shot capabilities of models trained on large robotics datasets~\cite{rt1, rt2, rtx, octo, roboagent, rth, rum, openvla, pi0}, but the curation of such datasets incurs a significant cost.
Some have exploited recent advances in VLMs, trained on ``web'' data without any connection to robotics, and directly elicit zero-shot robotic actions~\cite{codeaspolicies, copa, pivot, llmtrajgenerator, malmm, rekep, iker}. \jsadd{These policies are} 
limited by the lack of physical understanding and slow inference speed of VLMs, 
\axadd{as demonstrated by our experiments in Section~\ref{sec:comp_0shot}}.
Human web videos~\cite{sthsth, epickitchens, 100doh, howto100m, ego4d, egoexo4d}, due to their abundance, diversity, and rich information about interactions, emerge as a promising source of data for robotic skill acquisition. 

Since generating robot policies from out-of-domain human videos directly is difficult, many works instead train representations~\cite{r3m, hindex, hrp} (e.g. R3M~\cite{r3m}), rewards~\cite{dvd, vip, liv, progressor2024} (e.g. VIP~\cite{vip}), or affordances~\cite{mimicplay, screwmimic, whirl, egomimic, graphirl2022, atm2023, im2flow2act, general-flow, motiontracks, p3-po, hudor, pointpolicy, vidm, affordancediffusion, diffhoi, getagrip, ram2024, handsonvlm, star, vrb, swim, videodex, hopman, track2act, gen2act, hop2024, lapa} (See Fig~\ref{fig:related_work_map}). 
\jsadd{Some works~\cite{mimicplay, screwmimic, whirl, egomimic, graphirl2022, atm2023, im2flow2act, general-flow, motiontracks, p3-po, hudor, pointpolicy} (e.g. MimicPlay~\cite{mimicplay}, WHIRL~\cite{whirl}, and ATM~\cite{atm2023}) explored learning afoordances from in-domain human videos. Recent works~\cite{vidm, affordancediffusion, diffhoi, getagrip, ram2024, handsonvlm, star, vrb, swim, videodex, hopman, track2act, gen2act, hop2024, lapa} (e.g. VRB~\cite{vrb}, Track2Act~\cite{track2act}, LAPA~\cite{lapa}) extended these approaches to learning from in-the-wild human videos.}
Since these \jsadd{visual representations, reward functions, and} affordances are not explicitly actionable for robots, they still depend on \jdadd{in-domain} robot data to learn manipulation policies.

\jsadd{Very limited prior work~\cite{ditto, dinobot, orion, r+x, okami2024, vlmimic} such as DITTO~\cite{ditto}, R+X~\cite{r+x} and OKAMI~\cite{okami2024} has aimed to directly generate policies from human videos without any in-domain data.} 
These methods typically require the distribution of human demonstrations to be similar enough to the test-time robot environment and assumes knowledge of ground truth camera and depth information, making them unsuitable for learning from diverse and unstructured web data. 
Some methods also rely on heuristic-based mappings from human hand poses to robot gripper actions during data collection~\cite{r+x} or have manually defined constraint formulations~\cite{vlmimic}, limiting the range of demonstrations and tasks these methods can handle.

As Figure~\ref{fig:related_work_map} shows, to our knowledge, the only prior work that attempts zero-shot policies
from truly in-the-wild videos is H2R \cite{bharadhwaj2023zeroshot}, which learns plausible 3D hand trajectories from egocentric in-the-wild EpicKitchens~\cite{epickitchens} videos and retarget them to robot end effector for zero-shot deployment in real-world settings. 
We too train policies on EpicKitchens data, but with critical pre-processing steps that ground the data in 3D and generate higher quality policies. Further, we design a robust system that combines learned pre-contact interaction affordances and learned post-contact action policies.
As our experiments show, these method improvements translate to dramatic gains in the ability to generate functional out-of-the-box performance for manipulation skills in the real world. 

\section{Method}

\begin{figure*}[t]
\centering 
\includegraphics[width=0.88\textwidth]{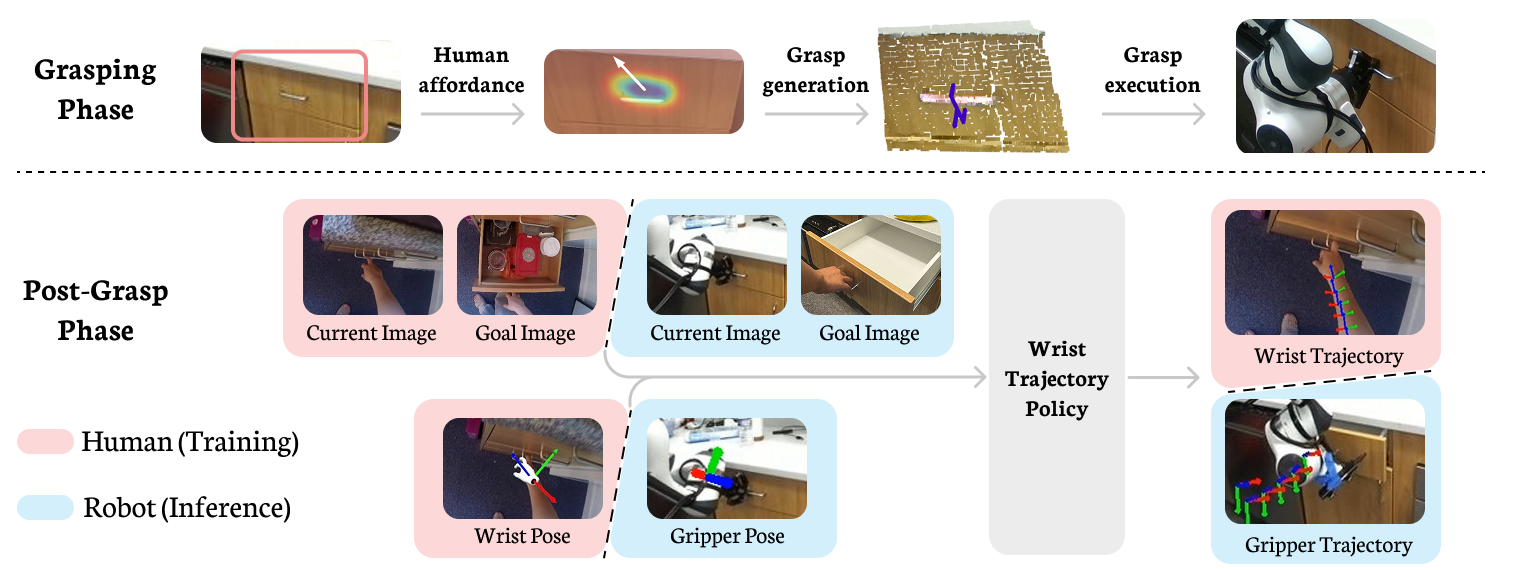}
\caption{\textbf{\methodname} is composed of the \textbf{grasping \jsadd{phase}} and the \textbf{post-grasp \jsadd{phase}}. The \textbf{grasping phase} (top)
\jsadd{leverages human affordance-based grasping}
to execute a task-relevant grasp.  The \textbf{post-grasp phase} (bottom) is an imitation policy trained on web videos to predict 6D wrist trajectories. We deploy this trained model directly on the robot.
}
\label{fig:method}
\vspace{-5mm}
\end{figure*}

\jdadd{We focus on manipulation skills that permit decomposition into two phases: the \textbf{grasping phase}  that consists of approaching and grasping an object of interest appropriately for the target task, and the \textbf{post-grasp phase}
which consists of a rigid manipulation of the object while stably held in the gripper. This encompasses such diverse skills as pick\&place, \axadd{slide opening and closing}, hinge opening and closing, pouring, cutting, and stirring.} \jdadd{\methodname pretrains components specific to these two phases, as described in {Sec~\ref{sec:grasping} and~\ref{sec:train-trajectory-model}} before combining them, as in Figure~\ref{fig:method}.} \jsadd{We focus on distilling human videos from EpicKitchens~\cite{epickitchens} into robotic skills.}
\jdadd{Pre-training on off-the-shelf human data naturally constrains our approach to be not tied to any specific robotic system design: we target static robot arms with 2-fingered grippers, observing the scene with an RGB-D camera from any egocentric-like vantage point of the robot workspace. See Appendix
\ref{app: setup}
for images and more details of our experiment setups.  
} 

\subsection{Human Affordance-Based Grasping}\label{sec:grasping} 

\jdadd{For this phase, we use human videos to learn to identify the appropriate region of the scene to seek to execute a grasp in, i.e., affordance prediction. Subsequent to this, given that human videos are of limited use in selecting the grasp itself due the vastly different embodiment of the robot's gripper and the human hand, we use} \jsadd{an approach trained on robot data to identify suitable} \jdadd{grasps for a 2-fingered gripper within that region, i.e. grasp selection.}

\jdadd{For affordance prediction, we use VRB~\cite{vrb} to generate a 3D point of intended contact. VRB is pre-trained on \jsadd{EpicKitchens~\cite{epickitchens}}. 
It generates pixel-space grasp locations, given an RGB image and a task description in natural language, e.g. ``open drawer''. Next, to select a grasp close to this chosen location, we use AnyGrasp~\cite{anygrasp}, a widely used grasp generation model pre-trained} \jsadd{on robot data} \jdadd{for our 2-fingered robotic grippers. Once a grasp is chosen, we plan a linear end-effector motion through free space to execute it.
See Figure \ref{fig:method} for examples of intermediate outputs after each stage of processing above, and the resulting grasp execution. 
}

\subsection{Human Movement-Based Post-Grasp Robot Policy}
\label{sec:train-trajectory-model}

Once the robot has grasped the object, it must decide what 6D end-effector trajectory to execute to accomplish the task. \methodname's post-grasp module is an imitation policy that distills this information from in-the-wild human videos. We first extract human wrist trajectories grounded in world 3D coordinates by reconstructing the hand pose and the egocentric camera, 
Given a skill, we take the corresponding subset of the data and train a skill model to predict 6D wrist trajectory. 

\paragraph{Extracting Human Wrist Trajectories From Web Videos} To curate diverse and large-scale human behavior, we use EpicKitchens~\cite{epickitchens}, an in-the-wild egocentric vision dataset. It contains 20M frames in 100 hours of daily activities in the kitchen.
To extract wrist trajectories from EpicKitchens, we run HaMeR~\cite{hamer}, a state-of-the-art pre-trained hand-tracking model, to obtain 3D hand pose reconstruction. HaMeR outputs the locations and orientations of all hand joints relative to a canonical hand, along with camera parameters corresponding to a translation \(t\in \mathbb{R}^3\).
\jdadd{We use camera parameters inferred through the COLMAP~\cite{colmap} structure-from-motion algorithm, as provided in EPIC-Fields~\cite{epicfields}, to convert these pixel-coordinate-based hand pose outputs into world 3-D coordinates.}
We consider only the wrist joint, and the result is 6D wrist trajectories \(\{h_t = (x_t, y_t, z_t, \alpha_t, \beta_t, \gamma_t)\}_{t=1}^{T}\) for a \(T\)-frame clip that is expressed in the world coordinate. See \href{https://zeromimic.github.io}{our website} for \jsadd{videos of} \methodname's hand reconstructions. 

\begin{figure}[h]
    \centering
    \begin{subfigure}[b]{0.45\columnwidth}
        \centering
        \includegraphics[width=\columnwidth]{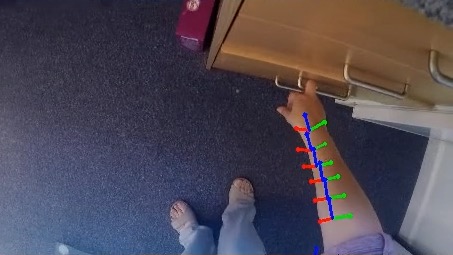}
        \caption{Open drawer}
        \label{fig:HAT_open_drawer}
    \end{subfigure}
    \hfill
    \begin{subfigure}[b]{0.45\columnwidth}
        \centering        \includegraphics[width=\columnwidth]{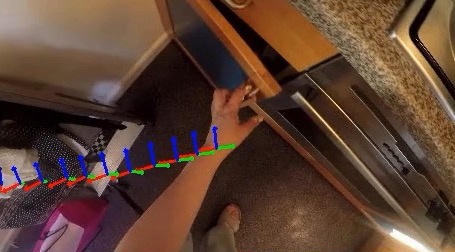}
        \caption{open cupboard}
        \label{fig:HAT_open_cupboard}
    \end{subfigure}
    \caption{6D wrist post-grasp policy outputs on unseen images. The red, green, and blue arrows denote the $x, y, z$ coordinates of the wrist orientation in the camera frame.}
    \label{fig:human_video_eval}
\vspace{-5mm}
\end{figure}

\paragraph{Policy Training, Execution, and Implementation Details}
A major challenge for learning to predict trajectories from web videos is the highly multi-modal nature of human demonstrations -- there are multiple ways to manipulate objects in a scene given the same image observation. To model this multi-modality, we use the recently popular action chunking transformer (ACT) \cite{aloha} policy class to learn a generative model over action sequences. The input of our model is the current image $I_t$, goal image $I_g$, and the current wrist pose $h_t$, and our model outputs future wrist poses $\{h_{i}\}_{i=t+i}^{t+n+1}$, where $n$ is the prediction chunk size. We use the last frame in the task-relevant clip as the goal image. See Figure \ref{fig:method} for an illustration. Since at test time, we perform robot experiments with \jsadd{a static camera}, we relieve the burden of the model to predict camera parameters by transforming all current and future wrist poses into the current frame's camera coordinate using the camera extrinsics of each frame. See Figure \ref{fig:human_video_eval} for qualitative visualization of generated wrist trajectories on unseen human videos. We train one model for each skill, obtaining a set of \numskills skill policies. Our model predicts relative 6D wrist poses with a chunk size of $n=10$, and we discuss the impact of these choices in Section \ref{sec:ablation}. We train each skill policy for 1000 epochs, which takes approximately 18 hours on an NVIDIA RTX 3090 GPU.

\paragraph{Retargeting Human Wrist Policy to the Robot} We deploy our trained post-grasp policies directly on the robot to generate 6D gripper trajectories (See Figure \ref{fig:method}). We use a single image of a human achieving the desired outcome as the ``goal image" for all trials of a task. In addition to the goal image, we provide the policy with the current
RGB observation and the current gripper pose in the camera's frame. The model predicts 6D trajectories in the same camera frame, which is converted to the robot frame for execution. The robot executes all actions in a chunk before prompting the model for the next round of inference.

\section{Experiments}

\begin{figure*}[t]
\centering 
\includegraphics[width=1.0\textwidth]{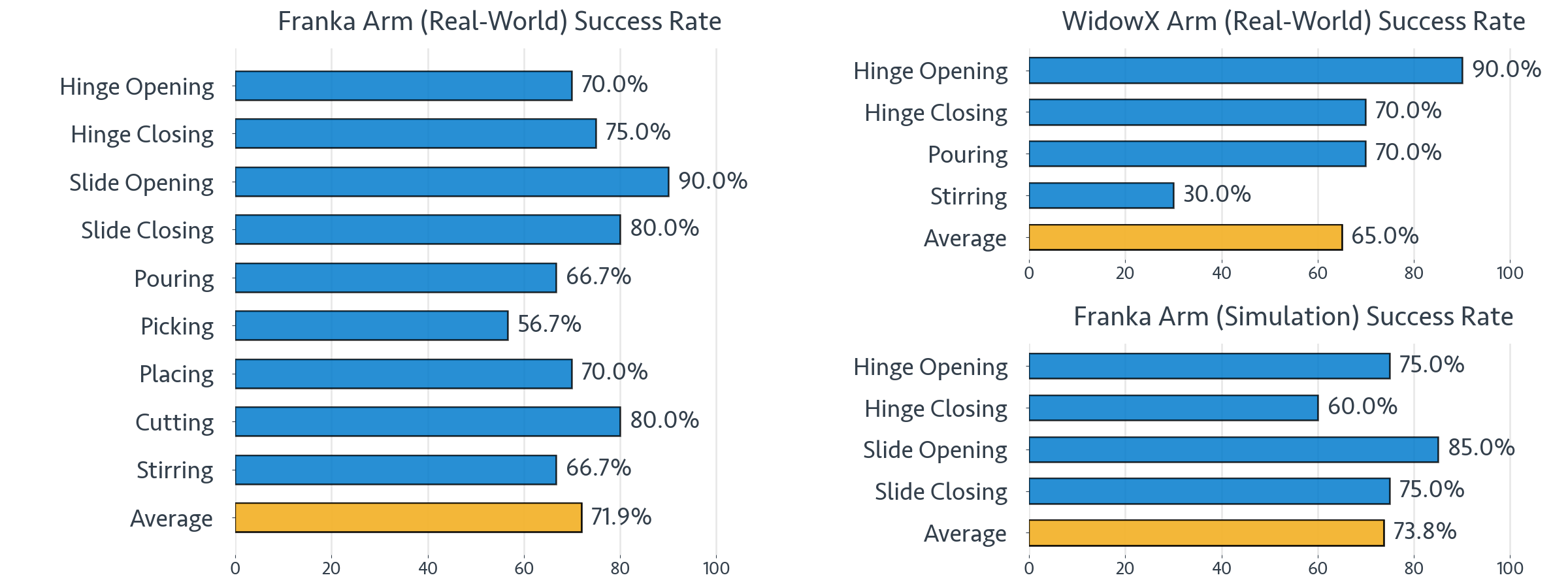}
\caption{\textbf{\methodname Zero-Shot Performance Overview}. \methodname demonstrates strong generalization capabilities, achieving consistent success across diverse tasks, robot embodiments, and both real-world and simulated environments.  The evaluation spans \numtotalinstances distinct scenarios across \numobjects object categories in 7 kitchen scenes, highlighting the adaptability and robustness of the system. For a detailed breakdown of performance by skills, robots, object categories, and scenarios, refer to Appendix~\ref{app:full_results}.
}
\label{fig:main_results}
\vspace{-5mm}
\end{figure*}

We evaluate \methodname skill policies out-of-the-box on a diverse set of real-world \axadd{and simulation} objects \axadd{with two different robot embodiments}. See our website \href{https://zeromimic.github.io}{\textbf{zeromimic.github.io}} for videos. Our experiments aim to answer the following questions:
\begin{enumerate}
    \item How important is each component of \methodname to its eventual performance?
    \item How well do \methodname skill policies perform when deployed zero-shot to perform varied skills in diverse real-world and \axadd{simulation} environments?
    \item \jsadd{How does \methodname compare to other state-of-the-art zero-shot robotic system?}
    \item What are the failure modes and causes of \methodname?
\end{enumerate}

\subsection{Experiment Setup}

\jsadd{We use the text annotations of EpicKitchens~\cite{epickitchens} to curate human video data corresponding to each manipulation skill.} Having trained the \numskills \methodname skill policies on in-the-wild human videos, we evaluate them on our robot in real-world \axadd{and simulation} environments. \axadd{Our real-world evaluation spans \numrealinstances distinct scenarios across \numobjects object categories in \numrealkitchenscenes kitchen scenes. In simulation, we evaluate 4 skill policies, randomizing kitchen scenes across trials.}
\axadd{Figure~\ref{fig:main_results} show the results.} \axadd{See
Appendix~\ref{app:full_results}
for a detailed breakdown of skills, robots, object categories, scenarios, and success rates.} None of the object instances or scenarios used in our experiments feature in our training data.

\paragraph{\axadd{Real-world experiment setup}} All \axadd{real-world} experiments are performed in \axadd{3 different} real kitchens \jdadd{on the UPenn campus} \axadd{and two different robots, a Franka Emika Panda arm and a Trossen Robotics WidowX 250 S arm.} Before our evaluations, we position the camera and the robot at a camera angle roughly similar to the relative camera angle of the human hand appearing in egocentric videos: camera at human height, and gripper within human arm's reach of the camera. We perform 10 trials with varying camera and robot positions to generally resemble a human's egocentric viewpoint. \axadd{The success of all experiments is determined by their consistency with the goal provided by the human goal image.} Visualizations of the trial positions are available on \href{https://zeromimic.github.io}{our website} \axadd{through an experiment time-lapse}. \axadd{See Appendix~\ref{app:real_setup}
for more details of our real-world experiment setup and images of our real kitchen scenes.}

\paragraph{\axadd{Simulation experiment setup}} \axadd{We conduct our simulation experiments in RoboCasa~\cite{robocasa}. We evaluate 4 \methodname skill policies, each across 20 randomized kitchen trials. For each trial, we vary the camera and robot positions, background objects, and kitchen styles (e.g., textures, object placements). We select camera views most similar to a human egocentric perspective.} \axadd{See Appendix~\ref{app:sim_setup}
for more details of our simulation experiment setup and images of our simulated kitchen scenes.}

\vspace{-1mm}
\subsection{Contribution of Each System Component to \methodname}
\label{sec:ablation}

We first validate the design of the \methodname procedure by measuring the importance of each of its components on two \axadd{real-world} tasks \axadd{with the Franka robot}: \textit{Open Drawer}  and \textit{Open Cupboard}. To do this, we construct ablated variants of \methodname that either drop a component or replace it with simpler alternatives. \jsadd{More details about the setup of these variants can be found in Appendix~\ref{app:ablation_details}. 
}

\paragraph{Grasping Methods} 
\methodname employs the human interaction affordance provided VRB \cite{vrb} to select which grasp produced by AnyGrasp \cite{anygrasp} to execute. We compare our approach to two simpler alternatives: (1) selecting the best grasp directly using AnyGrasp's grasp score (\textbf{Ours w/o interaction affordance}),
as done in \cite{okrobot}, and (2) moving the end effector to the 2D contact point lifted to 3D with depth, and close the gripper (\textbf{Ours w/o grasp model}), as done in \cite{vrb, swim}. The results in Table \ref{tab:grasp_baselines} indicate that \textbf{ours} is clearly the best method.
\textbf{Ours w/o interaction affordance} fails by proposing grasps on irrelevant scene regions, while \textbf{Ours w/o grasp model} struggles due to incorrect gripper orientations and imprecise contact predictions.

\begin{table}[!ht]
\vspace{-2mm}
\small
    \centering
    \begin{tabular}{cc>{\centering\arraybackslash}m{2cm}>{\centering\arraybackslash}m{2cm}}
        \toprule
        \textbf{Grasping Task} & \textbf{Ours} & \textbf{Ours w/o \newline Affordance} &\textbf{Ours w/o \newline Grasp Model} \\
        \midrule
        Drawer Handle & \textbf{8/10} &  0/10 &  0/10 \\
        Cupboard Handle & \textbf{7/10} &  4/10 &  6/10 \\
        \bottomrule
    \end{tabular}
    \caption{Success rates for different grasping methods.} 
    \label{tab:grasp_baselines}
\vspace{-3mm}
\end{table}

\paragraph{Wrist Post-Grasp Policy} After grasping the object, we deploy our 6D post-grasp policy to execute the task. H2R \cite{bharadhwaj2023zeroshot} also trains 6D wrist post-grasp policy on web videos, however the key difference is that it does not account for the impact of camera motion on the human hand motions detected in the video frames. We consider a strengthened version of H2R (\textbf{ours w/o SfM}) by simply removing camera extrinsics and intrinsics when processing our training data. Next, \textbf{VRB} is trained on web videos to produce post-contact trajectories only in terms of 2D pixel locations on the image, rather than 6-DOF wrist trajectories. To execute it on the robot, we sample a target end-point depth at random \jsadd{and interpolate the trajectory while fixing the gripper orientation}.

We evaluate the task success rate of our model and two alternatives after a successful grasp, and the results in Table \ref{tab:trajectory_baselines} show the superiority of our model, highlighting the importance of camera information from SfM and predicting dimensions beyond pixel coordinates. Both compared methods in this paragraph were designed as ablations of the post-contact wrist trajectory component of \methodname; as such, they benefit from \methodname's robust grasping phase. Without this, they would struggle still further: H2R cannot execute grasps in the original paper, and VRB does not provide grasp orientation even though it generates a contact point.

\begin{table}[h]
\small
    \centering
    \begin{tabular}{cccc}
        \toprule
        \textbf{Task} & \textbf{Ours} & \textbf{Ours w/o SfM} & \textbf{VRB} \\
        \midrule
        Open drawer & \textbf{10/10} & 4/10 & 2/10 \\
        Open cupboard & \textbf{10/10} & 6/10 & 0/10 \\
        \bottomrule
    \end{tabular}
    \caption{Success rates for different post-grasp policies \axadd{after a successful grasp.}}
    \vspace{-0.3cm}
    \label{tab:trajectory_baselines}
\end{table}

\jsadd{Additionally, to understand critical factors for predicting wrist trajectories from web videos, we evaluate several design choices of the post-grasp module using teleoperated successful grasps. We find that ACT~\cite{aloha} and Diffusion Policy~\cite{diffusionpolicy} policy architectures yield similar performance. Regarding action representation, relative actions in both translation and orientation significantly outperform absolute representations. Detailed results of these experiments are provided in Appendix~\ref{app:post_grasp_ablation}.
}
\vspace{-1mm}

\subsection{\methodname Zero-Shot Deployment Performance}

Having established the robustness of \methodname's system design above, we proceed to evaluate all \numskills \methodname skill policies zero-shot in varied real-world \axadd{and simulated} scenes with diverse objects and viewpoints. They achieve an impressive overall success rate of \axadd{\overallfrankasuccessrate in the real word with the Franka arm, \overallwidowxsuccessrate in the real world with the WidowX arm, and \overallsimsuccessrate in simulation}. See Figure~\ref{fig:main_results}
for a breakdown of success rates by skills. These results indicate that \methodname is capable of distilling a diverse set of unique skills from web videos. The results are best viewed in the videos presented on \href{https://zeromimic.github.io}{our website}.

The slide closing/opening and hinge closing/opening skills require grasping the object handle and reasoning about the object articulation affordances. Articulated objects often have handles of different shapes, sizes, and orientations, which our grasping module needs to appropriately adjust to. Furthermore, slide and hinge skills require different movements with respect to the object's articulation axis: translation and rotation, respectively. Hinge skills in particular require the model to determine if an object should be manipulated clockwise or counterclockwise along the axis (e.g. the left and right door of a cupboard).  

For the picking and placing skills, \methodname needs to reason about the target object pose provided in the goal image. Picking has the elevated complexity of grasping the object first, resulting in worse performance than the placing skill.

\methodname is also able to learn to use tools and perform pouring and cutting skills at a high level. Pouring requires reasoning about the target pour location and subsequently moving towards the location while rotating the object along the correct axis. Similarly, cutting requires reasoning about the cutting angle on the object given the initial knife pose and the target object pose. Afterwards, the robot needs to rotate the knife to align the edge and the object at the optimal angle and perform a swift downward motion. Interestingly, we observe that instead of always cutting straight down, which may result in an undesired cut on the object (e.g. slicing a vertically placed banana along its longer axis), our model is aware of the relative placements between the knife and the object and it learns to adjust its motion plan properly. 

Stirring is arguably the hardest skill to learn since it requires a particular set of motions where the translational position remains roughly the same but the orientation continuously moves in the same direction. Also, there is not much information about the desired motions in the goal image. In evaluation, \methodname can rotate a ladle and stir solid food objects as well as liquid in a container without excessive translational movements. 

\axadd{Having been trained exclusively on in-the-wild human videos, \methodname demonstrates remarkable generalization when deployed across object instances, categories, scenes, and robot embodiments. Notably, it successfully executes tasks involving object categories unseen in the human training data, such as \textit{Pour Salt from Spoon into Pan} and \textit{Cut Cake}. \methodname skill policies perform comparably on the WidowX and Franka arms for most tasks, except for the stirring skill, which is challenging due to the limited workspace of the smaller WidowX robot. Additionally, \methodname exhibits robustness to the real-to-sim gap, with no significant performance differences observed between real-world and simulation experiments.
}

\begin{table}[h]
\small
    \centering
    \setlength{\tabcolsep}{3pt} 
    \begin{tabular}{ccc}
        \toprule
        \textbf{Task} & \textbf{\methodname} & \textbf{ReKep \cite{rekep}} \\
        \midrule
        Open Drawer & \textbf{8/10} & 0/10 \\
        Close Drawer & \textbf{6/10} & \textbf{6/10} \\
        Place Pasta Bag into Drawer & \textbf{8/10} & 4/10 \\
        Pour Food from Bowl into Pan & \textbf{8/10} & 0/10 \\
        \bottomrule
    \end{tabular}
    \caption{\axadd{Success rates for different tasks using \methodname and ReKep.}}
    \vspace{-0.3cm}
    \label{tab:rekep}
\end{table}

\subsection{Comparison to Other Zero-Shot Robotic System}
\label{sec:comp_0shot}

\axadd{Concurrent work ReKep \cite{rekep} optimizes keypoint-based constraints generated by vision-language models (VLMs) to achieve zero-shot robotic behavior. Similar to \methodname, it does not require task-specific training or environment models. To compare \methodname to ReKep, we perform real-world experiments on 4 tasks using the Franka robot 
in a kitchen environment
(Figure~\ref{fig:levine_kitchen}).
The \textit{Open Drawer} and \textit{Close Drawer} tasks involve reasoning about the drawer's movement and articulation. The \textit{Place Pasta Bag into Drawer} task requires spatial reasoning to understand the relationships between objects. The \textit{Pour Food from Bowl into Pan} task demands reasoning about both object rotation and spatial relations.}

\axadd{Table \ref{tab:rekep} show the results. We observe that the failure cases of ReKep mostly stem from two issues: the vision module generates inaccurate keypoints or associates incorrect keypoints with target objects, and the VLM generates incorrect keypoint-based constraints due to its limited spatial reasoning capabilities. For more information about our implementation of ReKep and a detailed analysis of its failure cases, see
Appendix~\ref{app:rekep}.
}

\subsection{\methodname Failure Breakdown}

We investigate the system errors by examining the intermediate outputs of various modules and manually recording the cause of failure for each unsuccessful trial and aggregating their likelihood. \jsadd{Out of 87 failure trials in our real-world experiments, 31.1\% failed at the AnyGrasp stage, 24.1\% failed at the VRB stage, and 44.8\% failed at the post-grasp policy stage.} We present \jsadd{failure analysis of each module below and} several examples of these failures on \href{https://zeromimic.github.io}{our website}.

\jsadd{\textbf{AnyGrasp.} AnyGrasp is sensitive to point cloud sensing failures. We use the ``neural'' mode of Zed depth cameras for more accurate and smooth depth estimates; however, performance still degrades with small, reflective objects or under poor lighting conditions (e.g., small shiny drawer handles). Occasionally, AnyGrasp also generates incorrect or unreachable grasps.}

\jsadd{\textbf{VRB.}} A common issue with VRB is its difficulty in predicting appropriate contact locations \jsadd{on large furniture (e.g. cabinets, refrigerators) and} opened articulated objects. \jsadd{Additionally, since VRB internally relies on Grounded SAM~\cite{groundedsam} for language-based segmentation, segmentation errors can directly result in its failures.}

\jsadd{\textbf{Post-grasp policy}. The post-grasp policy is sometimes sensitive to camera-robot relative positional configurations, especially if they deviate significantly from an egocentric perspective, since the policy models are trained on egocentric human data. Additionally, action reconstructions from human videos are inherently noisy, causing difficulties with fine-grained tasks such as pouring from a spoon or cutting small food items.}

\section{CONCLUSIONS \& LIMITATIONS}

We have presented \methodname, a first step towards distilling zero-shot deployable a repertoire of robotic manipulation skill policies from purely in-the-wild human videos, each validated in real scenes with real objects. Presently, \methodname exploits a simplified pre-grasp / post-grasp skill stricture, directly retargets human wrist movements to the robot without accounting for morphological differences, does not learn any in-hand manipulations, \jsadd{non-prehensile interactions,} or gripper release, and does not handle tasks requiring two arms. Nevertheless, we have shown that it already suffices to learn many useful skills. \methodname builds on the very best current models and hardware for grasp generation, interaction affordance prediction, depth sensing, and hand detection. We have shown that it is limited by their performance; as those models continue to improve, they will further increase the viability of our approach. Finally, we have trained \methodname on a relatively modest 100 hrs of egocentric daily activity dataset, and expanding this to include larger datasets such as Ego-4D~\cite{ego4d} and beyond could help to generate a more comprehensive and performant repository of web-distilled skill policies.

\section*{\jsadd{ACKNOWLEDGMENT}}

\jsadd{This work is funded by NSF CAREER 2239301, NSF 2331783, DARPA TIAMAT HR00112490421, and UPenn University Research Fellowship.}






\clearpage



\clearpage
\appendices
\begin{table*}[ht]
\centering
\footnotesize
\begin{tabular}{ccccc}
    \multicolumn{5}{c}{\axadd{\textbf{Real-World Results}}} \\ 
    \midrule
    \textbf{Skill} & \axadd{\textbf{Robot}} & \textbf{Object Category} & \axadd{\textbf{Scenario}} & \textbf{Success Rate (\%)} \\
    \midrule
    \axdrop{Vertical }Hinge Opening 
    & Franka & Cupboard & \axadd{Levine Hall Kitchen} & 6/10 \\
    & \axadd{Franka} & \axadd{Cupboard} & \axadd{Table Top 1} & \axadd{6/10} \\
    & \axadd{Franka} & \axadd{Cupboard} & \axadd{Table Top 2} & \axadd{8/10} \\
    & \axadd{Franka} & \axadd{Fridge} & \axadd{GRASP Lab Kitchen} & \axadd{8/10} \\
    & \axadd{WidowX} & \axadd{Cupboard} & \axadd{Table Top 3} & \axadd{9/10}\\
    \midrule
    \axdrop{Vertical }Hinge Closing 
    & Franka & Cupboard & \axadd{Levine Hall Kitchen} & 4/10 \\
    & \axadd{Franka} & \axadd{Cupboard} & \axadd{Table Top 1} & \axadd{8/10} \\
    & \axadd{Franka} & \axadd{Cupboard} & \axadd{Table Top 2} & \axadd{10/10} \\
    & \axadd{Franka} & \axadd{Fridge} & \axadd{GRASP Lab Kitchen} & \axadd{8/10} \\
    & \axadd{WidowX} & \axadd{Cupboard} & \axadd{Table Top 3} & \axadd{7/10} \\
    \midrule
    Slide Opening 
    & Franka & Drawer & \axadd{Levine Hall Kitchen} & 8/10 \\
    & \axadd{Franka} & \axadd{Drawer} & \axadd{Towne Hall Kitchen} & \axadd{10/10} \\
    \midrule
    Slide Closing 
    & Franka & Drawer & \axadd{Levine Hall Kitchen} & 6/10 \\
    & \axadd{Franka} & \axadd{Drawer} & \axadd{Towne Hall Kitchen} & \axadd{10/10} \\
    \midrule
    Pouring 
    & Franka & Water from Bowl into Sink & \axadd{Levine Hall Kitchen} & 8/10 \\
    & Franka & Food from Bowl into Pan & \axadd{Levine Hall Kitchen} & 8/10 \\
    & \axadd{Franka} & \axadd{Salt from Spoon into Pan} & \axadd{Levine Hall Kitchen} & 4/10 \\
    & \axadd{WidowX} & \axadd{Water from Cup into Pot} & \axadd{Table Top 3} & \axadd{7/10} \\
    \midrule
    Picking 
    & Franka & Can & \axadd{Levine Hall Kitchen} & 7/10 \\
    & Franka & Banana & \axadd{Levine Hall Kitchen}n & 4/10 \\
    & \axadd{Franka} & \axadd{Marker} & \axadd{Table Top 1} & \axadd{6/10} \\
    \midrule
    Placing 
    & Franka & Spoon & \axadd{Levine Hall Kitchen} & 10/10 \\
    & Franka & Pasta Bag into Drawer & \axadd{Levine Hall Kitchen} & 4/10 \\
    \midrule
    Cutting 
    & Franka & Tofu & \axadd{Levine Hall Kitchen} & 8/10 \\
    & Franka & Banana & \axadd{Levine Hall Kitchen} & 8/10 \\
    & Franka & Cake & \axadd{Levine Hall Kitchen} & 8/10 \\
    \midrule
    Stirring 
    & Franka & Food in Pan & \axadd{Levine Hall Kitchen} & 6/10 \\
    & Franka & Pasta in Water & \axadd{Levine Hall Kitchen}& 8/10 \\
    & \axadd{Franka} & \axadd{Water in Pan} & \axadd{Table Top 1} & \axadd{6/10} \\
    & \axadd{WidowX} & \axadd{Food in Pot} & \axadd{Table Top 3} & \axadd{3/10} \\
    \midrule
    \addlinespace[1.5mm]
    \textbf{\numskills Skills} & \axadd{\textbf{\numrobots Robots}} & \axadd{\textbf{\numobjects Categories}} & \axadd{\textbf{\numrealinstances Total Instances}} & \axadd{\textbf{\overallrealsuccessrate}} \\ 
    \midrule
    \addlinespace[4.0mm]
    \multicolumn{5}{c}{\axadd{\textbf{Simulation Results}}} \\ 
    \midrule
    \axadd{\textbf{Skill}} & \axadd{\textbf{Robot}} & \axadd{\textbf{Object Category}} & \axadd{\textbf{Scenario}} & \axadd{\textbf{Success Rate (\%)}} \\
    \midrule
    \axadd{\axdrop{Vertical }Hinge Opening}
    & \axadd{Franka} & \axadd{Cupboard} & \axadd{Simulated Kitchen} & \axadd{15/20} \\
    \midrule
    \axadd{\axdrop{Vertical }Hinge Closing} 
    & \axadd{Franka} & \axadd{Cupboard} & \axadd{Simulated Kitchen} & \axadd{12/20} \\
    \axadd{Slide Opening}
    & \axadd{Franka} & \axadd{Drawer} & \axadd{Simulated Kitchen} & \axadd{17/20} \\
    \midrule
    \axadd{Slide Closing} 
    & \axadd{Franka} & \axadd{Drawer} & \axadd{Simulated Kitchen} & \axadd{15/20} \\
    \midrule
    \addlinespace[1.5mm]
    \axadd{\textbf{4 Skills}} & \axadd{\textbf{1 Robots}} & \axadd{\textbf{2 Categories}} & \axadd{\textbf{4 Total Instances}} & \axadd{\textbf{\overallsimsuccessrate}} \\ 
    \bottomrule
\end{tabular}
\caption{Summary of skills, \axadd{robots}, object categories, \axadd{scenarios}, and success rates for \axadd{real-world and simulation results.}}
\label{table:combined_results}
\vspace{-5mm}
\end{table*}

\section{Experimental Setup Details}
\label{app: setup}

\subsection{Real-World Experimental Setup Details}
\label{app:real_setup}

\begin{figure}[h]
\centering 
\includegraphics[width=0.6\columnwidth]{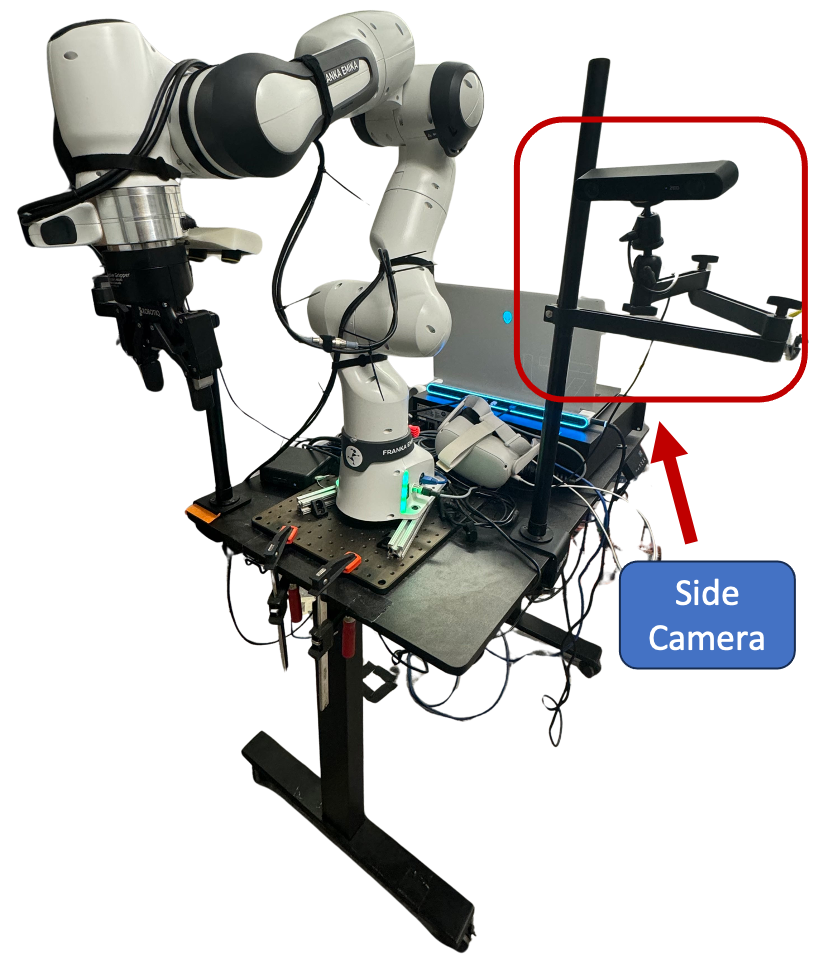}
\caption{Our Franka hardware setup includes a 7-DOF Franka Emika Panda arm with a Robotiq 2-fingered gripper and a Zed 2 stereo camera mounted on the base.}
\label{fig:franka_setup}
\end{figure}

\begin{figure}[h]
\centering 
\includegraphics[width=0.6\columnwidth]{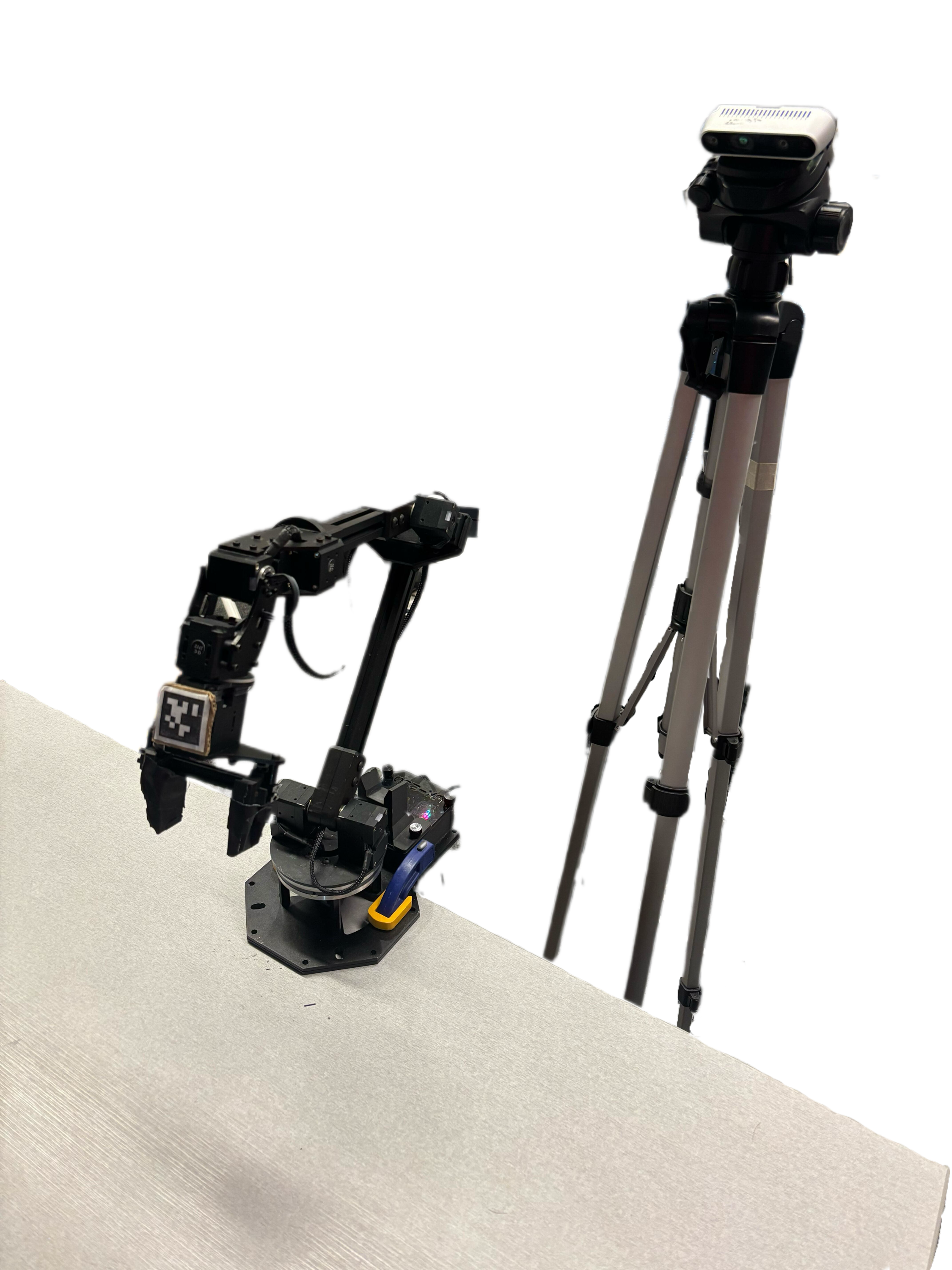}
\caption{\axadd{Our WidowX hardware setup includes a 6-DOF Trossen Robotics
WidowX 250 S arm attached to a table with a 2-fingered gripper and an Intel RealSense Depth Camera D435.}}
\label{fig:widowx_setup}
\end{figure}

\axadd{Our Franka experiments} uses the hardware setup in Fig~\ref{fig:franka_setup}, \axadd{which is similar to that used in prior works \cite{droid}}. We use a Franka Emika Panda arm with a Robotiq two-finger gripper mounted on a mobile base, which we use to drag the robot across various scenes. We use a Zed 2 stereo camera mounted on the base \axadd{to capture RGB and depth images}. 

\begin{figure}[h]
    \centering
    \begin{tabular}{cc}
        \subcaptionbox{Levine Hall Kitchen\label{fig:levine_kitchen}}{\includegraphics[width=0.48\columnwidth]{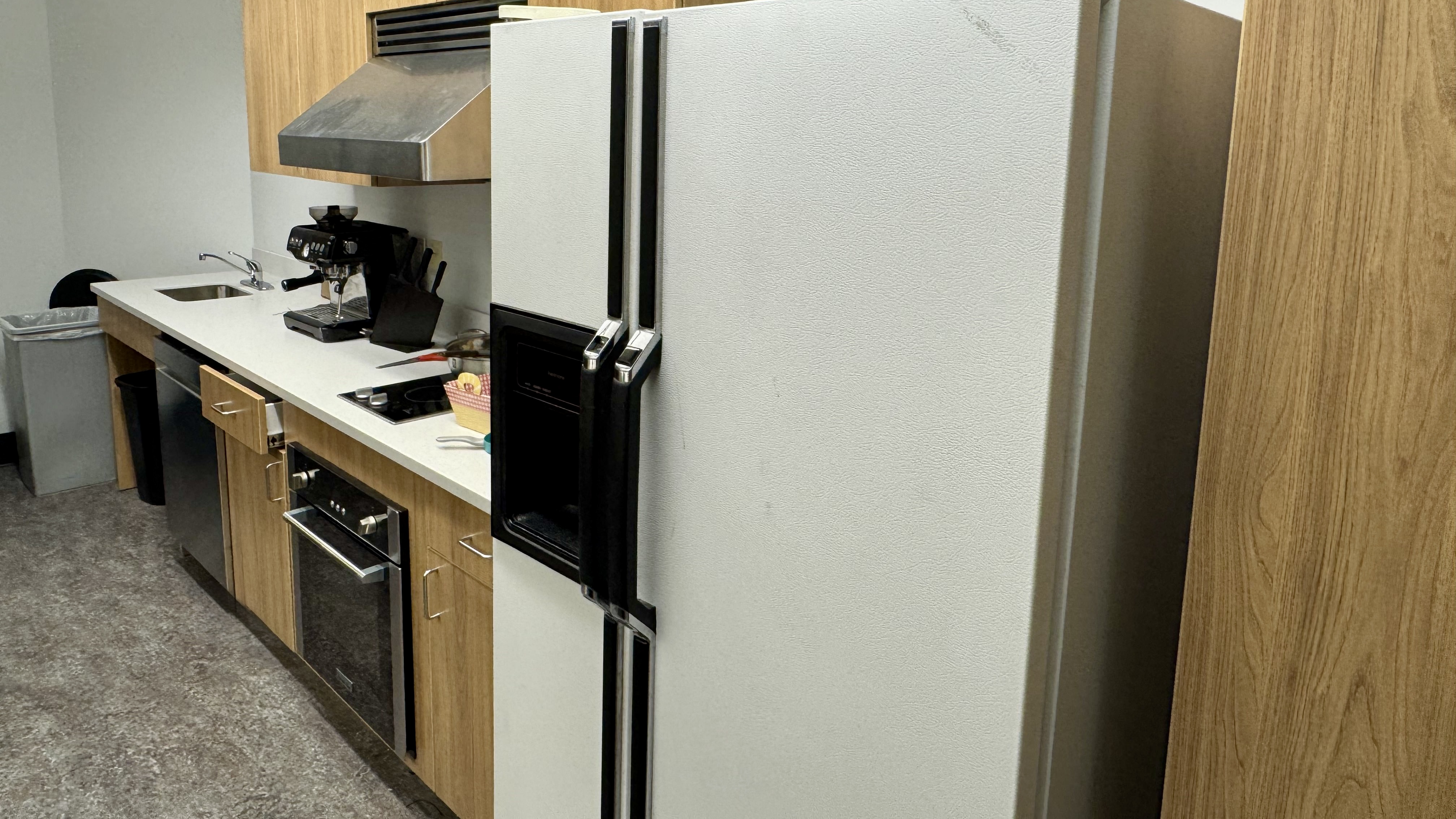}} &
        \subcaptionbox{Grasp Lab Kitchen\label{fig:grasp_kitchen}}{\includegraphics[width=0.48\columnwidth]{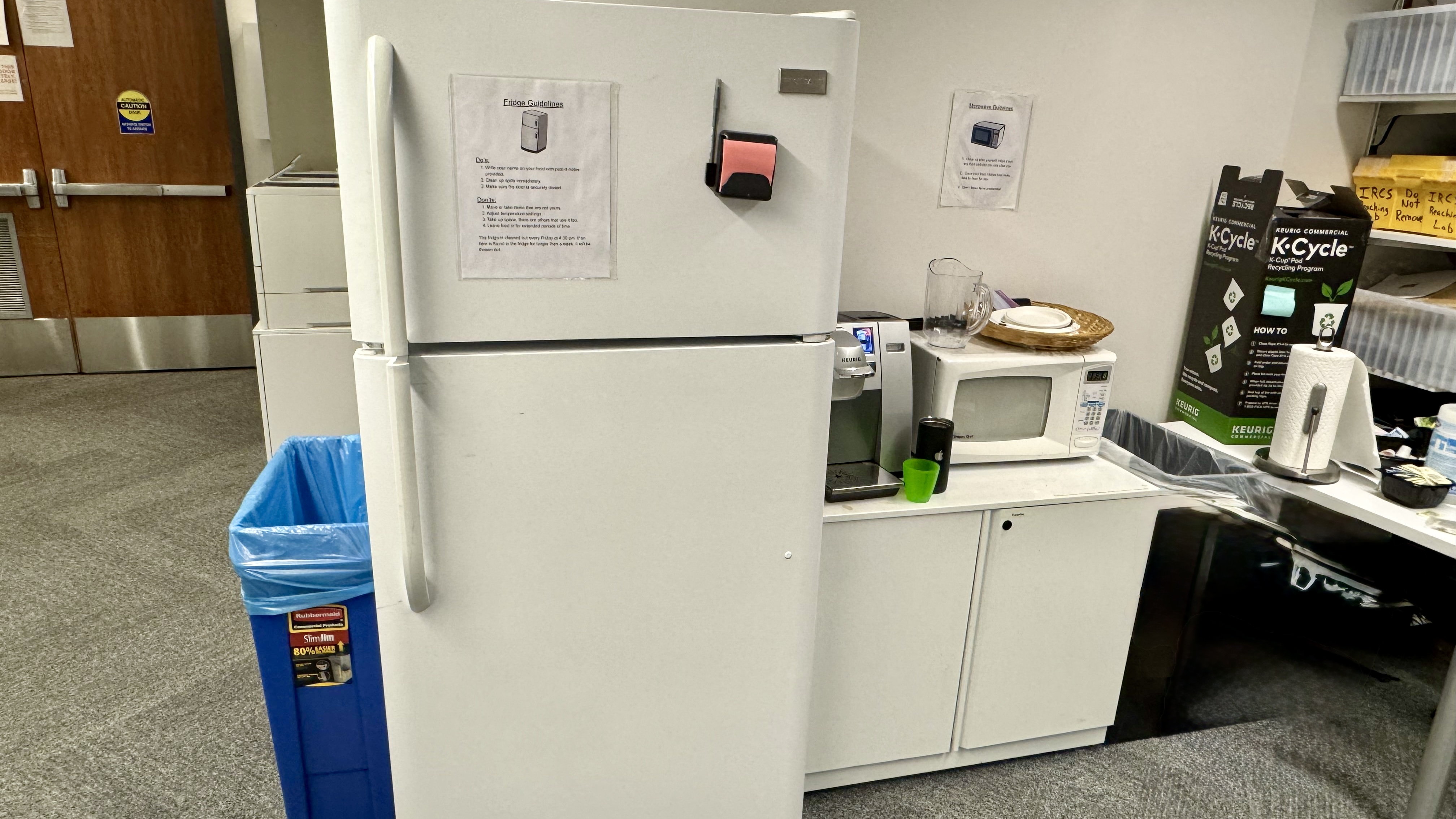}} \\
        \subcaptionbox{Towne Hall Kitchen\label{fig:towne_kitchen}}{\includegraphics[width=0.48\columnwidth]{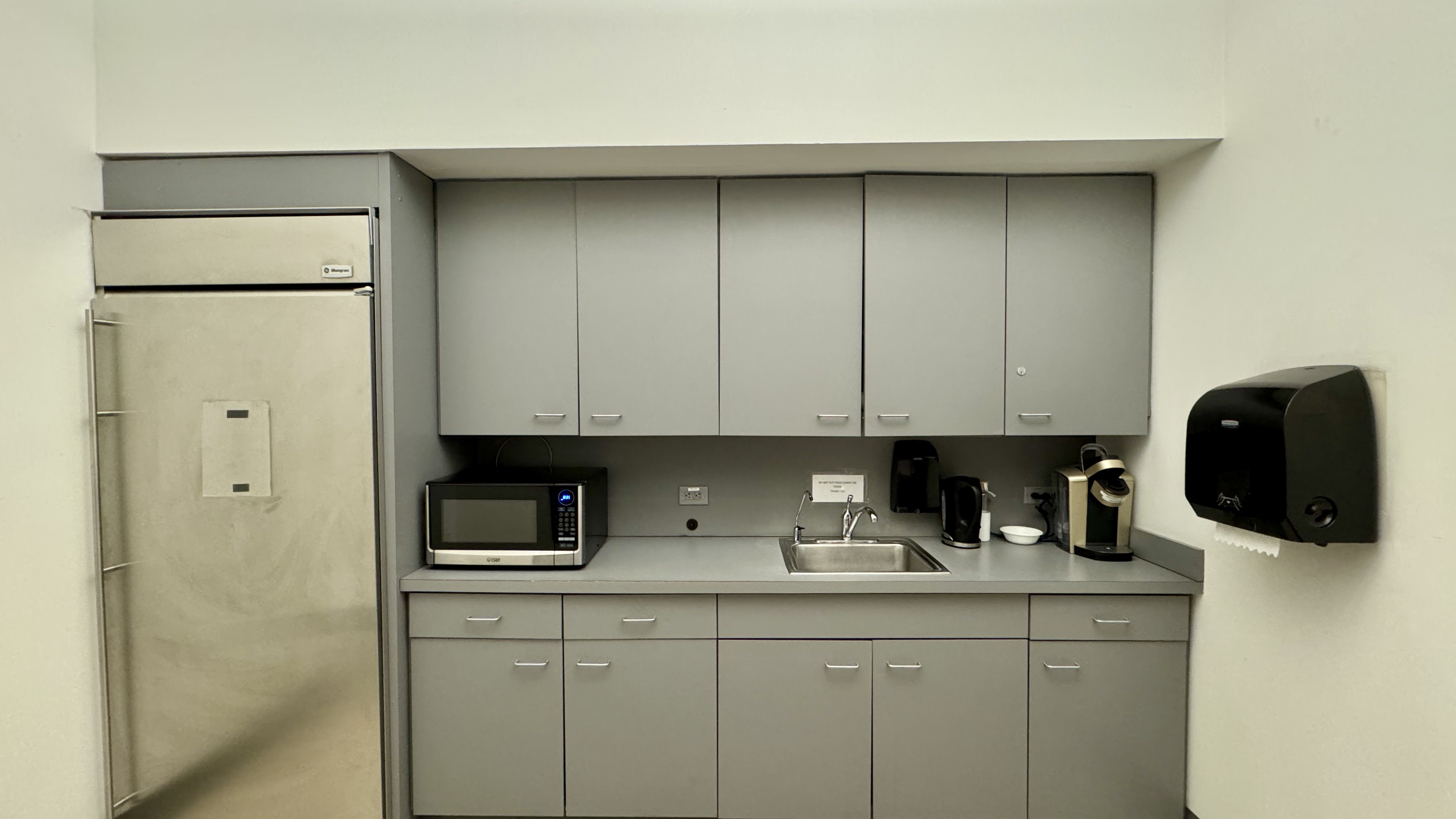}} &
        \subcaptionbox{Table Top 1\label{fig:table_top1}}{\includegraphics[width=0.48\columnwidth]{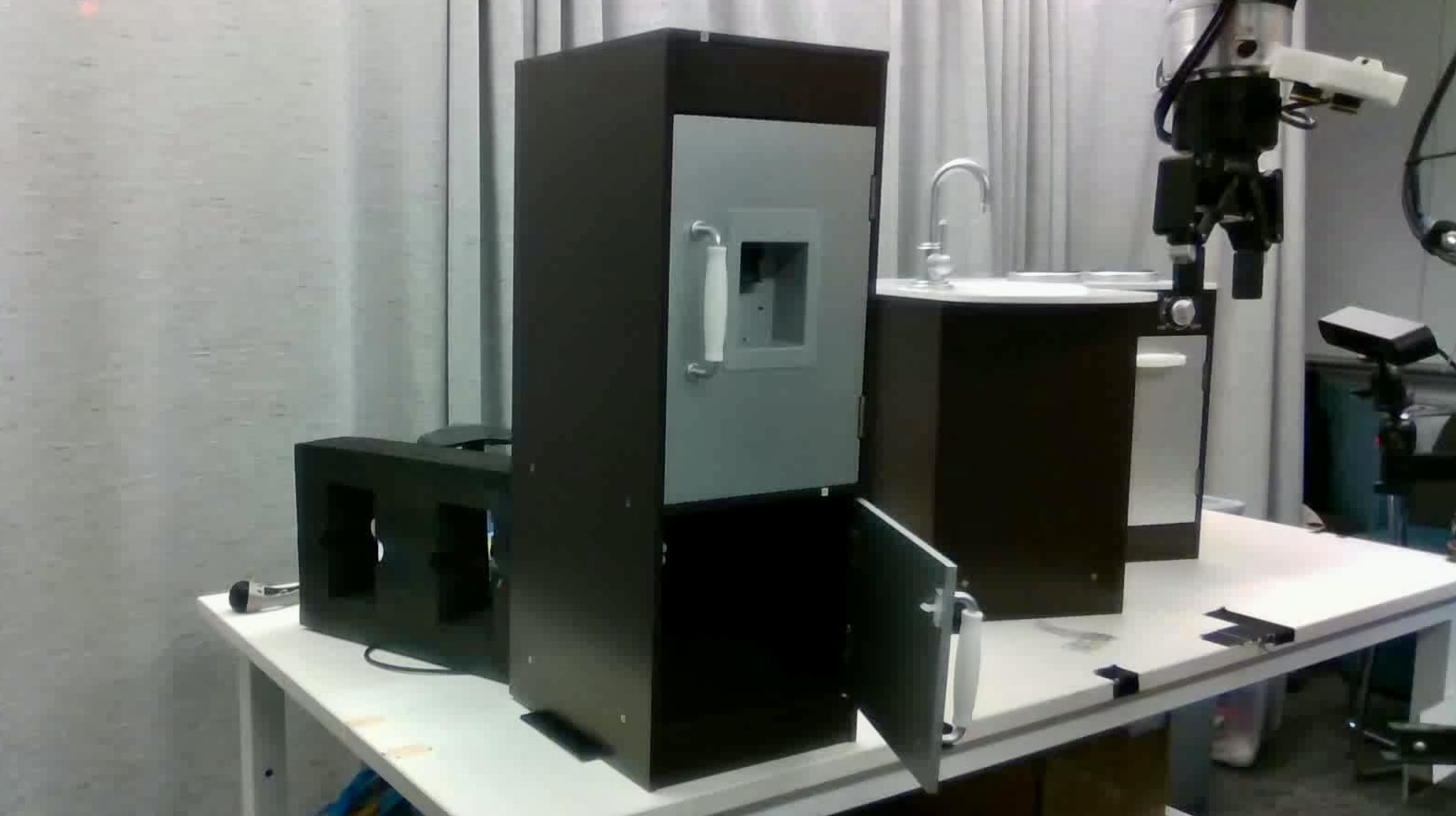}} \\
        \subcaptionbox{Table Top 2\label{fig:table_top2}}{\includegraphics[width=0.48\columnwidth]{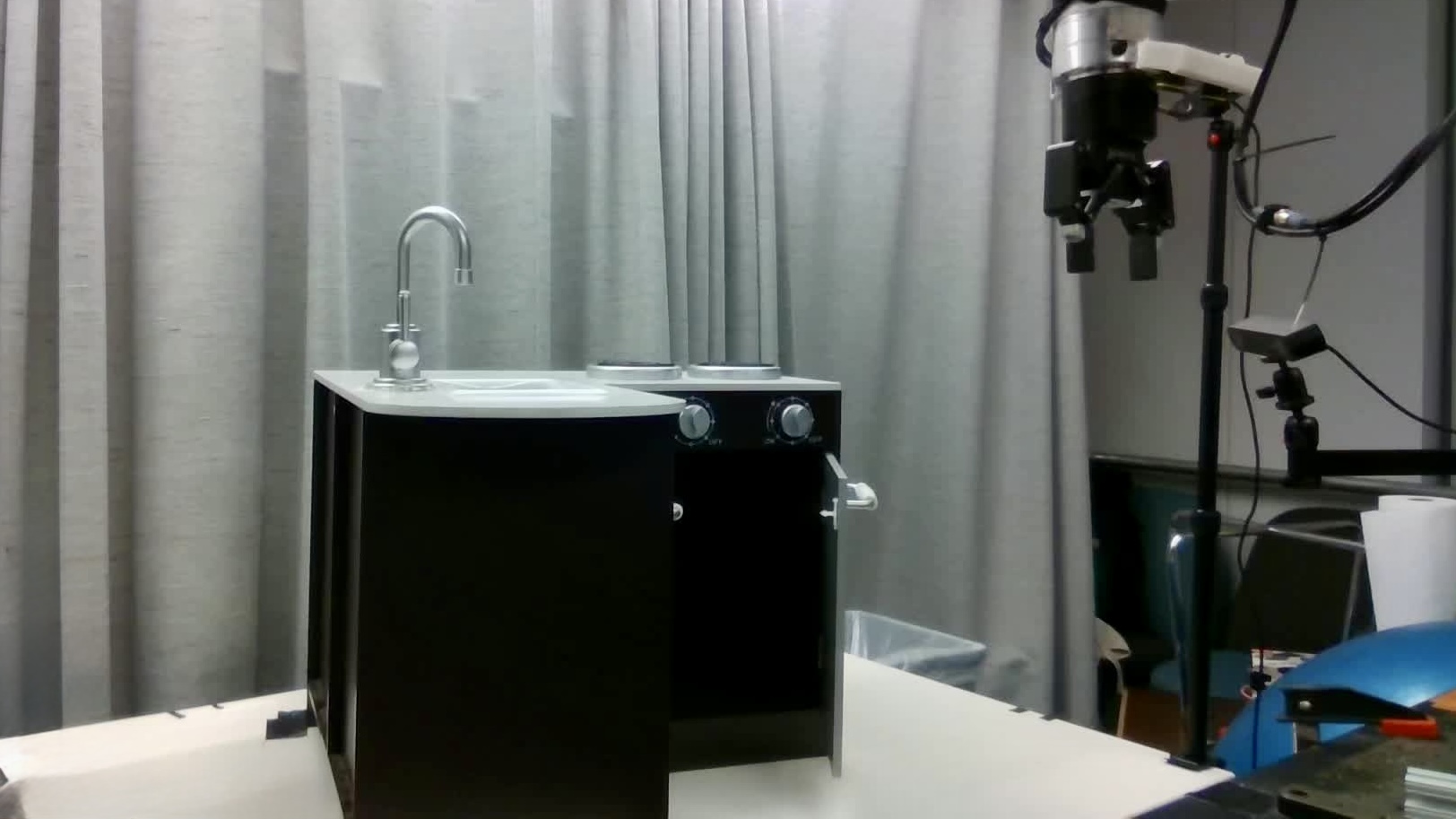}} &
        \subcaptionbox{Table Top 3\label{fig:table_top3}}{\includegraphics[width=0.48\columnwidth]{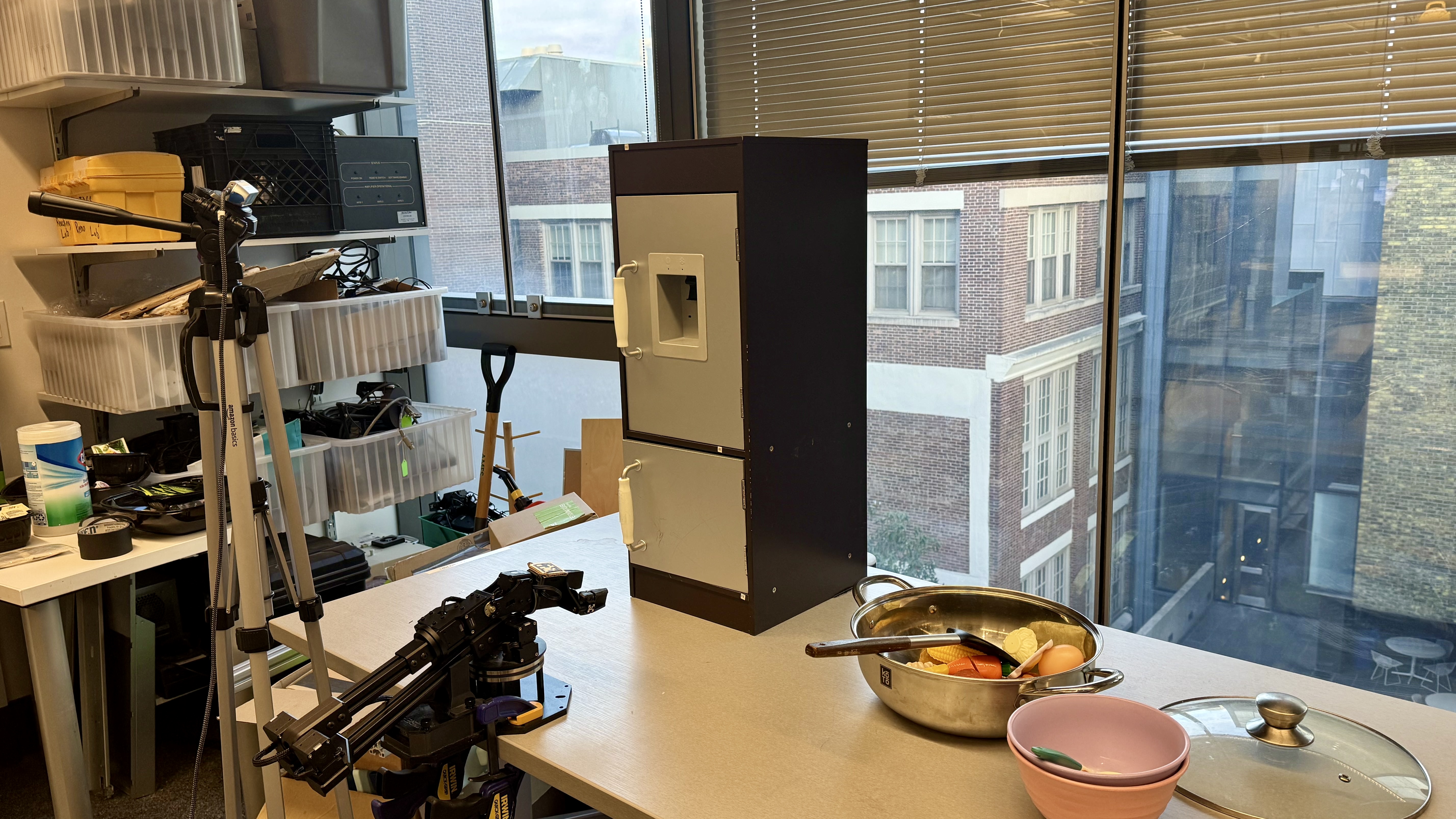}} \\
    \end{tabular}
    \caption{\axadd{Real-world environments used in our experiments: (a-c) Various kitchen environments across different buildings, (d-f) Different tabletop setups. We perform our Franka experiments using setups (a-e), and our WidowX experiments using setup (f).}}
    \label{fig:real_environments}
\end{figure}

\axadd{Our WidowX experiments uses the hardware setup in Fig~\ref{fig:widowx_setup}. The WidowX arm is attached to a stationary table. An Intel RealSense Depth Camera D435 is mounted on a tripod beside the table to capture RGB and depth images.}

\axadd{Figure \ref{fig:real_environments} shows our real-world experimental environments. We conducted experiments in three different kitchen environments (Figures \ref{fig:levine_kitchen}-\ref{fig:towne_kitchen}) and three tabletop setups (Figures \ref{fig:table_top1}-\ref{fig:table_top3}). For the Franka robot experiments, we used environments (a)-(e), moving the robot between different buildings. The WidowX robot experiments were conducted using the stationary tabletop setup shown in (f).}

\subsection{Simulation Experimental Setup Details}
\label{app:sim_setup}

\begin{figure}[h]
\centering 
\includegraphics[width=1.0\columnwidth]{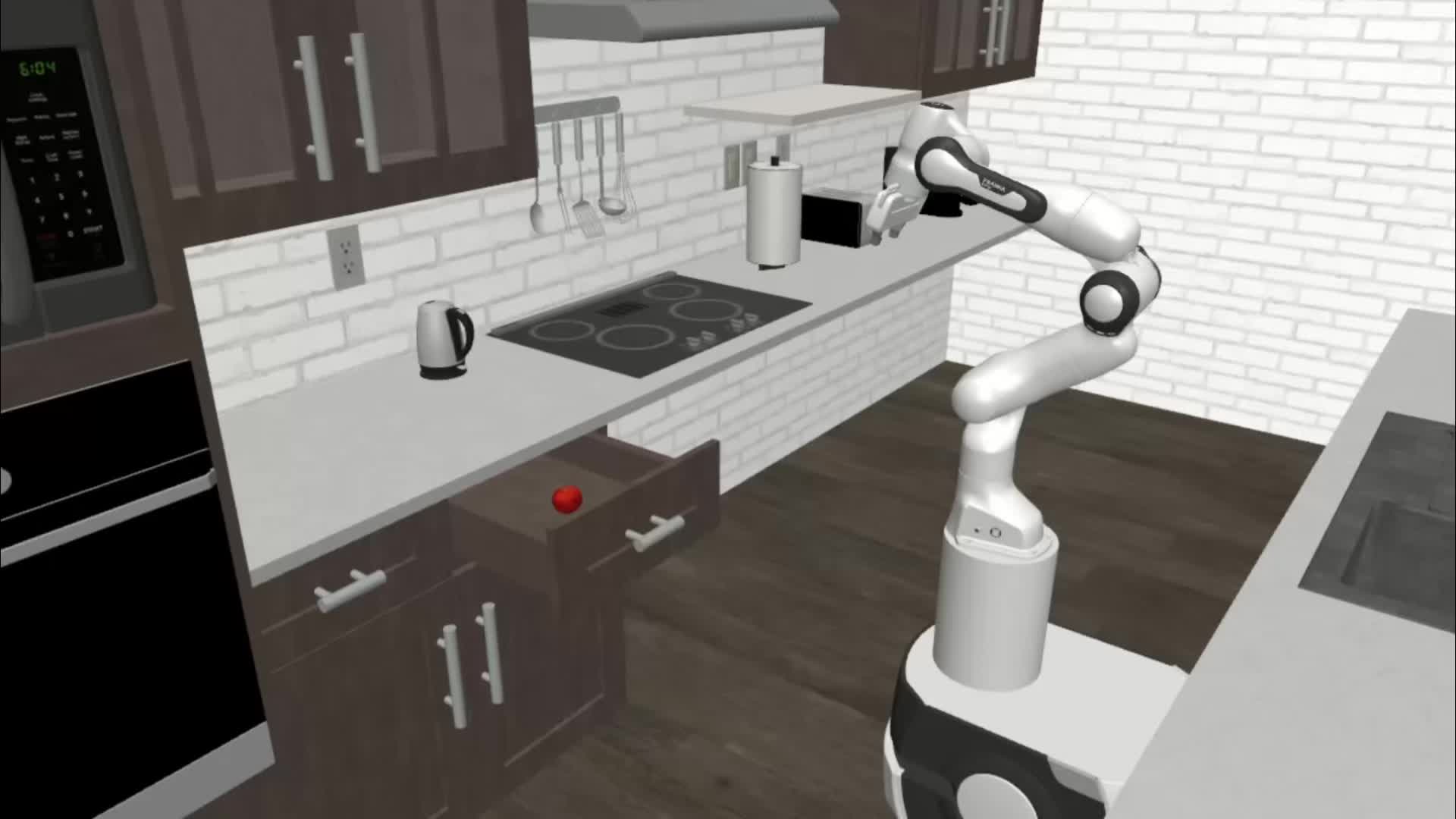}
\caption{\axadd{Our RoboCasa simulation setup includes a 7-DOF Franka Emika Panda arm with a 2-fingered gripper.}}
\label{fig:robocasa_setup}
\end{figure}

\begin{figure}[h]
\centering
\begin{tabular}{cc}
    \includegraphics[width=0.48\columnwidth]{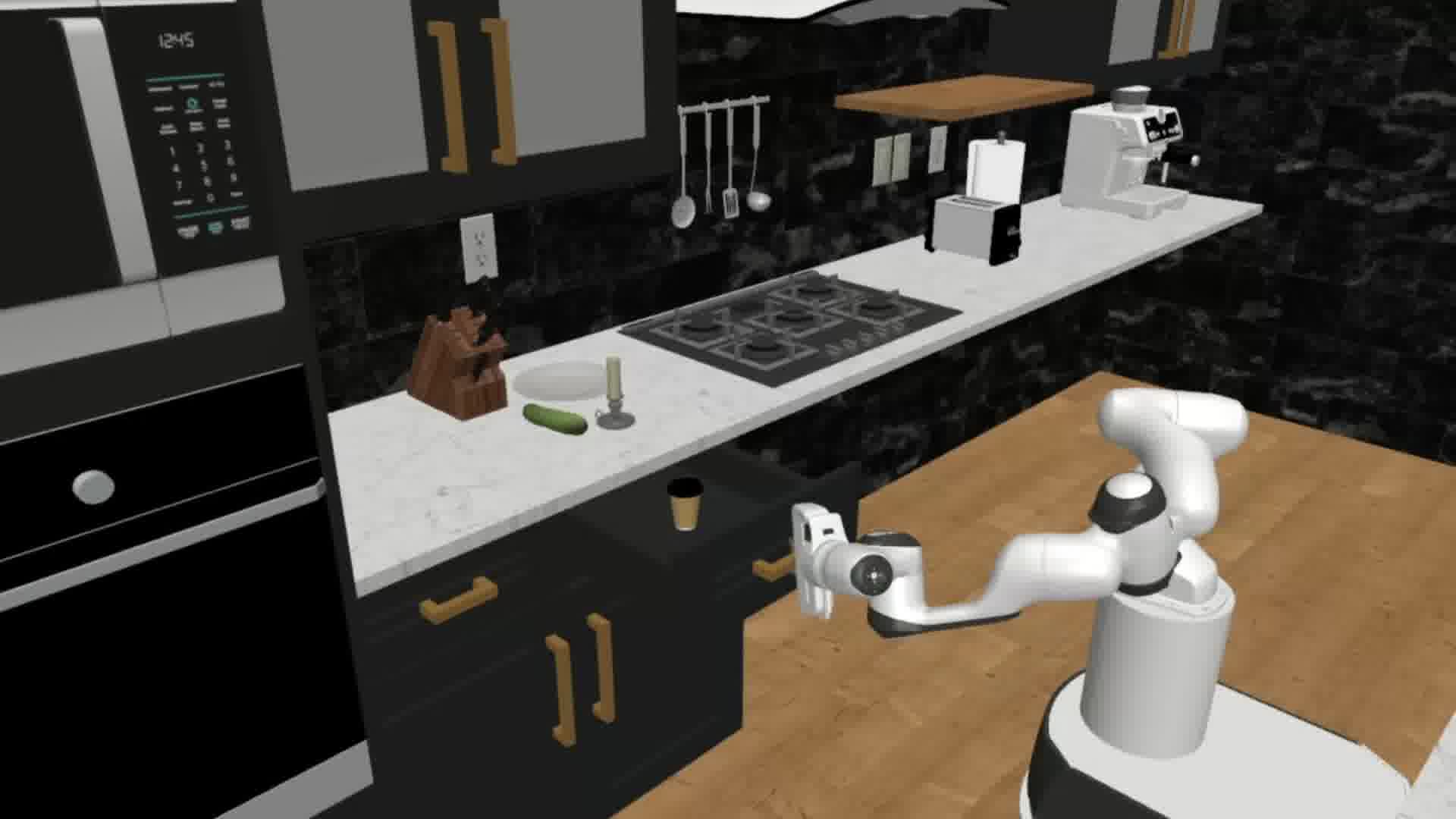} &
    \includegraphics[width=0.48\columnwidth]{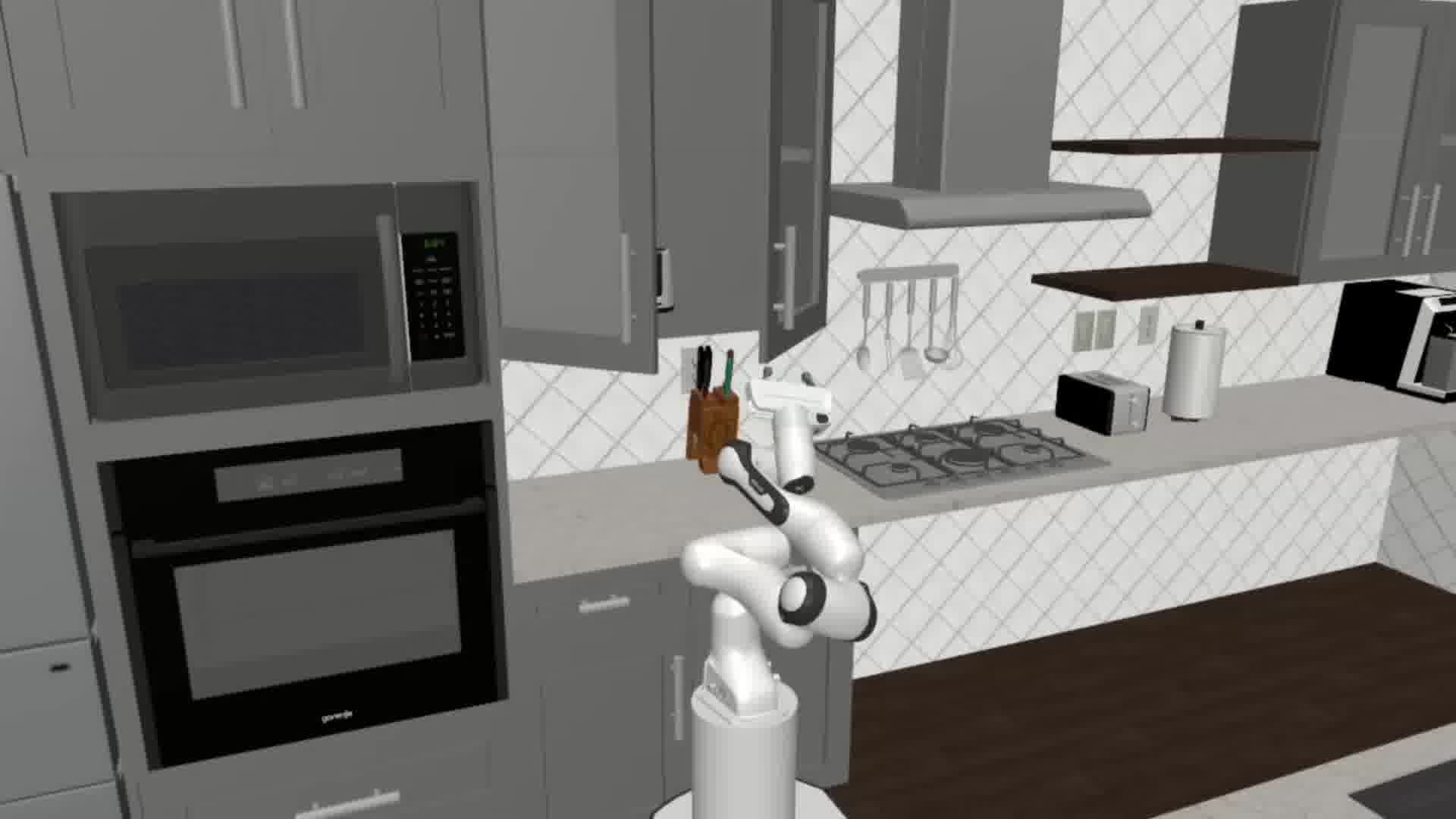} \\
    \includegraphics[width=0.48\columnwidth]{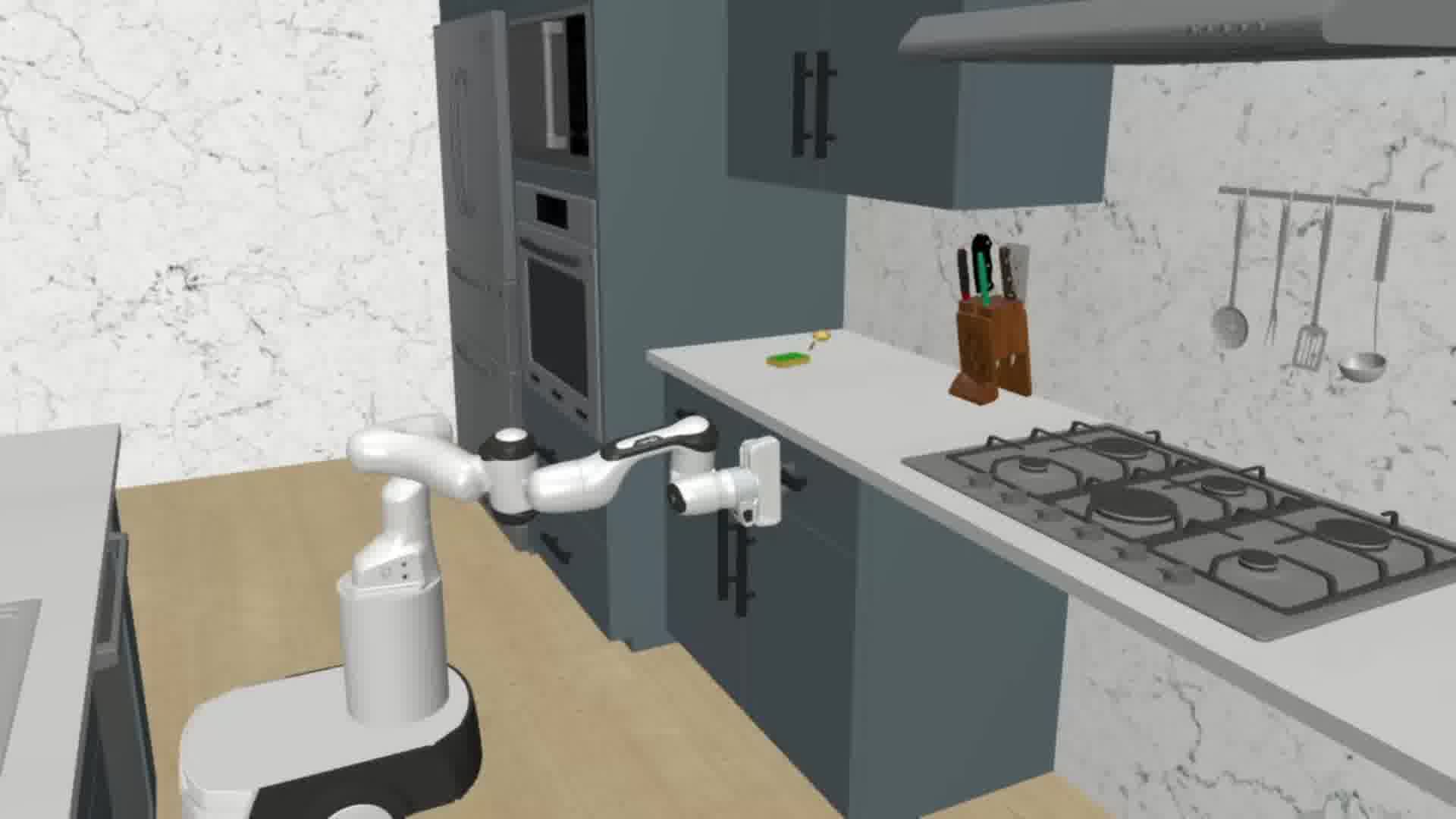} &
    \includegraphics[width=0.48\columnwidth]{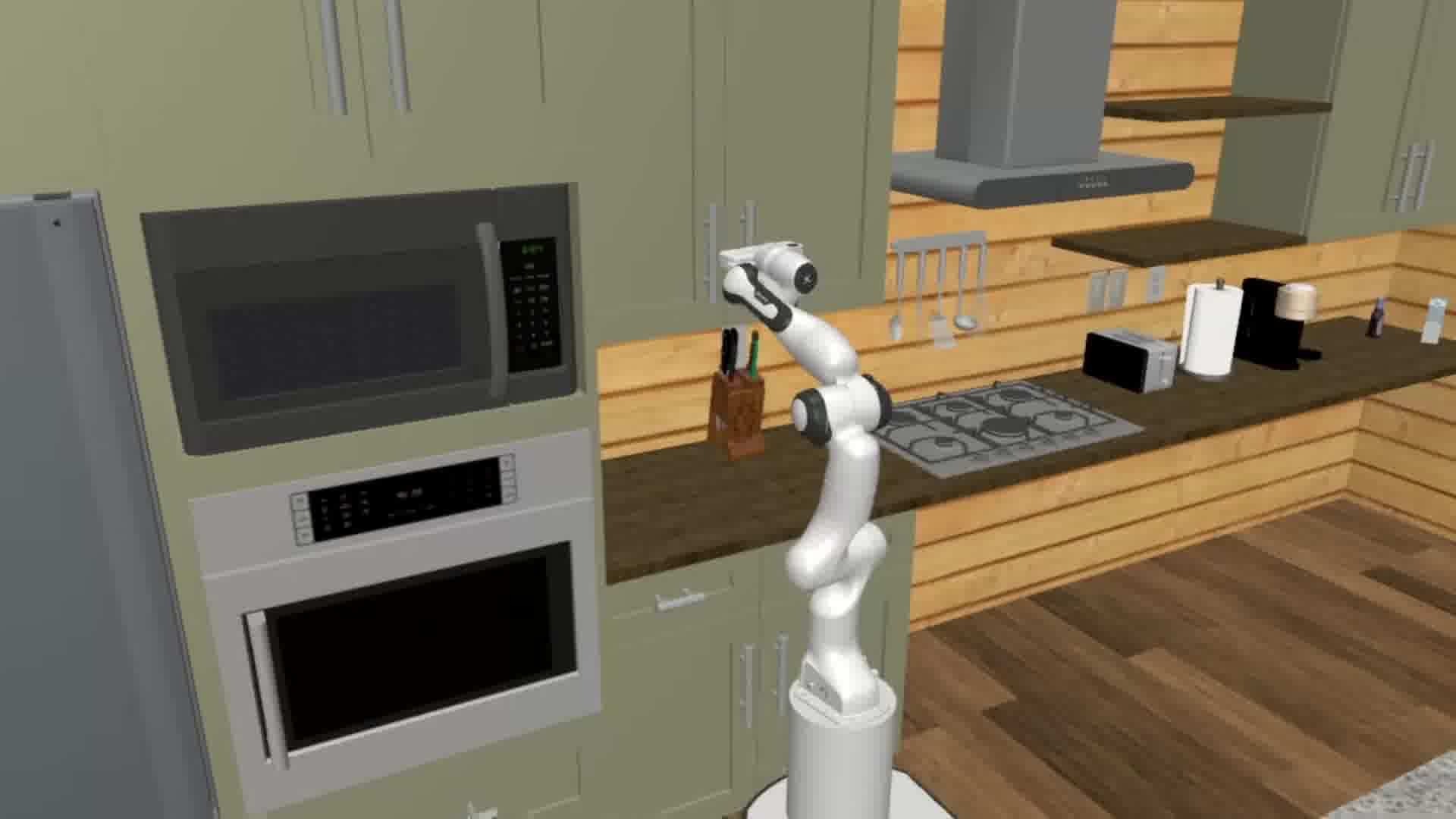} \\
\end{tabular}
\caption{\axadd{RoboCasa environment images showcasing different setups and configurations. Each image corresponds to a different kitchen environment style.}}
\label{fig:robocasa_environment}
\end{figure}

\axadd{Figure~\ref{fig:robocasa_setup} shows our simulation setup in RoboCasa \cite{robocasa}. We use a Franka Emika Panda arm with a two-finger gripper in a simulated kitchen layout. We perform 20 trials for each task, varying the camera position, randomizing the robot's position, altering the background object instances and their positions, and selecting a random kitchen style from the 12 available options. Each kitchen style features unique textures, distractor objects, and fixture attributes, such as cabinet and drawer handle types. Figure \ref{fig:robocasa_environment} are example images of different kitchen scene styles. The success of a trial is evaluated based on the specific success conditions defined for each task provided by RoboCasa.}

\section{Detailed Breakdown of \methodname Zero-Shot Deployment Performances}
\label{app:full_results}

\axadd{Table~\ref{table:combined_results} provides a comprehensive overview of \methodname's performance across both real-world and simulation environments. The results are categorized by skills, robots, object categories, and scenarios, offering insight into the system's versatility and adaptability.}

\axadd{In the real-world evaluation, we assessed \numskills skills across \numrobots robots and \numobjects object categories, spanning \numrealinstances distinct scenarios. These evaluations resulted in an overall success rate of \overallrealsuccessrate. Additionally, we evaluated 4 skills in a controlled simulated kitchen environment using one robot across two object categories, totaling four distinct scenarios. This simulation study achieved an overall success rate of \overallsimsuccessrate.}

\section{Ablation Experiment Details}
\label{app:ablation_details}

\jsadd{In Section~\ref{sec:ablation}, the \textit{Open Drawer} task is performed with the ``slide opening'' skill policy, and the \textit{Open Cupboard} task is performed with the ``hinge opening'' skill policy.
}
\subsection{Grasping Methods Ablation Details}
\label{app:grasp_ablation_details}
\textbf{Ours w/o grasp model} is an ablated variant of \methodname where the end effector is moved to the 2D contact point proposed by VRB~\cite{vrb} lifted to 3D with depth, and the gripper is then closed. In this experiment, since VRB does not output orientation, we use the gripper's initial orientation for the grasp pose. 

\subsection{Wrist Post-Grasp Policy Ablation}
\label{app:post_grasp_ablation_details}
\jsadd{In our strengthened version of H2R (\textbf{ours w/o SfM}), we remove camera extrinsics and intrinsics when processing our training data. As a result, the 3D location of the wrist is only represented by its pixel coordinate and hand size, the output of depth-ambiguous monocular hand reconstruction methods \cite{frankmocap, hamer}.}

\jsadd{\textbf{VRB} produces post-contact trajectories only in terms of 2D pixel locations on the image. To execute it on the robot, we convert VRB's 2D outputs to 6D using the following procedure: we sample a target end-point depth at random and interpolate the waypoints while fixing the gripper orientation as the initial post-grasp gripper orientation throughout the trajectory. 
}

\section{Additional Post-Grasp Module Ablation Experiments}
\label{app:post_grasp_ablation}

To gain insight into what is critical to learning to predict wrist trajectories from web videos, we teleoperate the robot to obtain a successful grasp and then evaluate a number of alternative post-grasp trajectory generation options and present our findings in this section.

\paragraph{Imitation Policy Architecture} We compare ACT \cite{aloha} and Diffusion Policy \cite{diffusionpolicy}, two popular imitation learning policy classes, for training our post-grasp policy on EpicKitchens. As illustrated in Table \ref{tab:training_algorithms}, they perform similarly \axadd{when evaluated in the real world with the Franka robot}. ACT performs slightly better on skills that mostly require gripper translation, while Diffusion Policy is marginally better at more rotation-heavy tasks. For consistency, we use ACT for all of our other experiments and ablations.

\begin{table}[!ht]
\vspace{-1mm}
    \small
    \centering
    \begin{tabular}{cccc}
        \toprule
        \textbf{Method} & \textbf{Open drawer} & \textbf{Open cupboard} & \textbf{Pour water} \\
        \midrule
        ACT & \textbf{10/10} & \textbf{8/10} & 7/10 \\
        DiffPo & 8/10 & \textbf{8/10} & \textbf{9/10} \\
        \bottomrule
    \end{tabular}
    \caption{Success rates for different post-grasp policies \axadd{after a successful grasp}.}
    \label{tab:training_algorithms}
\vspace{-3mm}
\end{table}

\paragraph{Relative vs. Absolute Action Representation} For both the translation (\textbf{T}) dimensions and the orientation (\textbf{O}) dimensions, we compare training an ACT model with absolute and relative action representations, resulting in four variants: absT+absO, absT+relO, relT+absO, and relT+relO. Evaluating on the \axadd{real-world} ``pour water" task \axadd{with the Franka arm}, their respective success rates are 1/10, 3/10, 2/10, and 7/10, indicating that relT+relO performs significantly better than other variants.
We hypothesize that the orientation distribution shift from the human hand to the gripper as well as discontinuity in orientation space from $-\pi$ to $\pi$ makes it harder for the model to learn meaningful absolute orientation representation.

\section{ReKep Baseline Details}
\label{app:rekep}

\subsection{ReKep Implementation Details}
\axadd{We adapted the publicly released simulation code of ReKep~\cite{rekep} for OmniGibson to integrate with our real-world Franka arm setup~\cite{droid}. To evaluate ReKep as a zero-shot system without human intervention, we use its "Auto" mode, which automatically generates keypoints and constraints, instead of the "Annotation" mode, which requires manual annotation for both. Our implementation is available at: \href{https://github.com/Everloom-129/ReKep}{https://github.com/Everloom-129/ReKep}.}

\axadd{As part of the adaptation, we rewrote the environment module, including the robot controller and keypoint registration components. To ensure optimal performance and to use ReKep's steelman version as a competitive baseline, we rely on teleoperated grasping for its grasping module, effectively minimizing grasp failures. In the perception module, we replace the ground-truth masks provided by the simulator with those generated by the Segment Anything Model 2 (SAM2) \cite{sam2}. These masks are filtered using area upper and lower bounds to ensure accuracy. Additionally, we modify ReKep's original k-means and mean-shift clustering algorithms to refine the generated keypoints, providing the VLM with cleaner input data for generating keypoint constraints. Lastly, we replace the simulator's ground-truth depth data with depth data from a ZED 2 Stereo Camera. We use its neural depth mode and apply band filtering to improve the accuracy and reliability of depth values.}

\subsection{ReKep Failure Cases}
\axadd{We present specific examples and a detailed analysis of ReKep's failure cases across three tasks with low success rates: \textit{Open Drawer}, \textit{Place Pasta Bag into Drawer}, and \textit{Pour Food from Bowl into Pan}.}

\begin{figure}[h]
\centering 
\includegraphics[width=1.0\columnwidth]{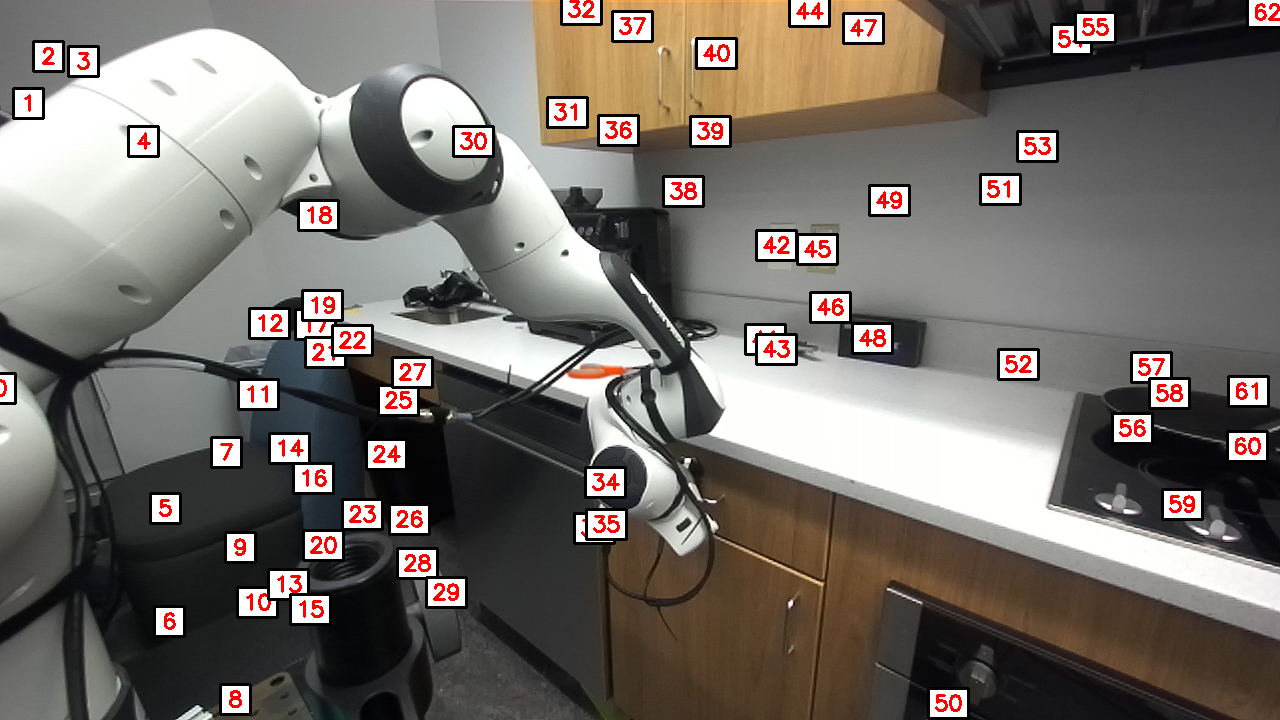}
\caption{\axadd{Keypoints proposed by ReKep for the \textit{Open Drawer} task.}}
\label{fig:open_drawer_rekep_keypoints}
\end{figure}

\begin{lstlisting}[style=pythoncode, caption=\axadd{The constraints generated by ReKep for the \textit{Open Drawer} task instruct the end effector to move leftward in the camera frame (Line 12). However, this direction deviates from the drawer's actual outward articulation axis.}, label=code:open_drawer_rekep_constraints]
def path_constraint1(end_effector, keypoints):
    """The robot must still be grasping the drawer 
    handle (keypoint 35)."""
    handle_position = keypoints[35]
    cost = np.linalg.norm(end_effector - handle_position)
    return cost
    
def subgoal_constraint1(end_effector, keypoints):
    """The drawer handle (keypoint 35) should be displaced 
    outward by 10cm along the x-axis."""
    handle_position = keypoints[35]
    offsetted_position = handle_position + np.array([-0.1, 0, 0])
    cost = np.linalg.norm(handle_position - offsetted_position)
    return cost
\end{lstlisting}

\paragraph{\textit{Open Drawer}} \axadd{Failures in the \textit{Open Drawer} task arise because the VLM struggles with identifying the drawer's articulation axis in the camera frame, causing the gripper to become stuck. Figure~\ref{fig:open_drawer_rekep_keypoints} illustrates the keypoints proposed by ReKep, while Code Snippet~\ref{code:open_drawer_rekep_constraints} presents the corresponding constraints generated by ReKep. These constraints direct the end effector to move 10 cm along the negative x-axis (leftward) in the camera frame. However, the actual outward articulation axis of the drawer corresponds to $-x$ (left), $+y$ (down), and $-z$ (towards the screen) in the camera frame. As a result, the actions generated by ReKep's constraints cause the gripper to become stuck, despite its attempts to move.}

\begin{figure}[h]
\centering 
\includegraphics[width=1.0\columnwidth]{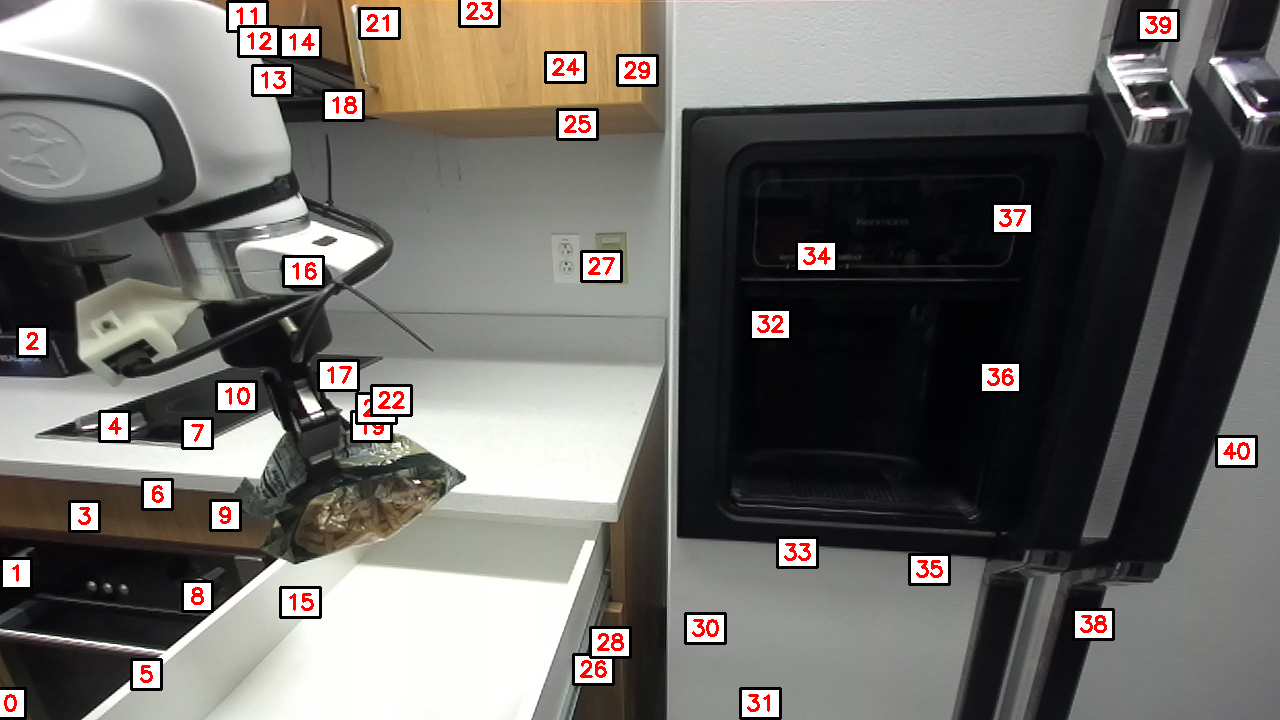}
\caption{\axadd{Keypoints proposed by ReKep for the \textit{Place Pasta Bag into Drawer} task.}}
\label{fig:place_pasta_rekep_keypoints}
\end{figure}

\begin{lstlisting}[style=pythoncode, caption=\axadd{For the \textit{Place Pasta Bag into Drawer} task, ReKep generates constraints based on incorrectly identified keypoints. Specifically, it misclassifies keypoint 22, a background keypoint, as the pasta keypoint, and keypoint 6, another background keypoint, as the drawer keypoint. See Lines 16-17 and Figure~\ref{fig:place_pasta_rekep_keypoints} for the misclassified keypoints.}, label=code:place_pasta_rekep_constraints]
def path_constraint1(end_effector, keypoints):
    """
    Ensure the robot is still grasping the pasta bag during the movement.
    The cost is the Euclidean distance between the end-effector and the pasta bag's keypoint (keypoint 22).
    """
    pasta_bag_keypoint = keypoints[22]
    cost = np.linalg.norm(end_effector - pasta_bag_keypoint)
    return cost
    
def subgoal_constraint1(end_effector, keypoints):
    """
    Ensure the pasta bag is inside the drawer.
    The cost is the Euclidean distance between the pasta bag's keypoint (keypoint 22) 
    and the drawer's keypoint (keypoint 6).
    """
    pasta_bag_keypoint = keypoints[22]
    drawer_keypoint = keypoints[6]
    cost = np.linalg.norm(pasta_bag_keypoint - drawer_keypoint)
    return cost
\end{lstlisting}

\paragraph{\textit{Place Pasta Bag into Drawer}}\axadd{Figure~\ref{fig:place_pasta_rekep_keypoints} illustrates the keypoints proposed by ReKep, while Code Snippet~\ref{code:place_pasta_rekep_constraints} presents the corresponding constraints. The keypoint proposal reveals that ReKep's vision module struggles to generate a reliable keypoint on the inside of an empty drawer. Additionally, ReKep projects 3D keypoints onto 2D images, which can result in spatially close keypoints overlapping and cause errors in the VLM's keypoint selection. For example, it identifies a keypoint near the edge of the pasta bag but slightly outside its actual boundary as belonging to the bag. This misplacement leads to the keypoint's depth value being incorrectly interpreted as the larger background depth value. Additionally, it sometimes associates nearby background keypoints with the drawer. By generating constraints based on these misidentified keypoints, ReKep produces ineffective movement instructions for the end effector, ultimately resulting in task failure.}

\begin{figure}[h]
\centering 
\includegraphics[width=1.0\columnwidth]{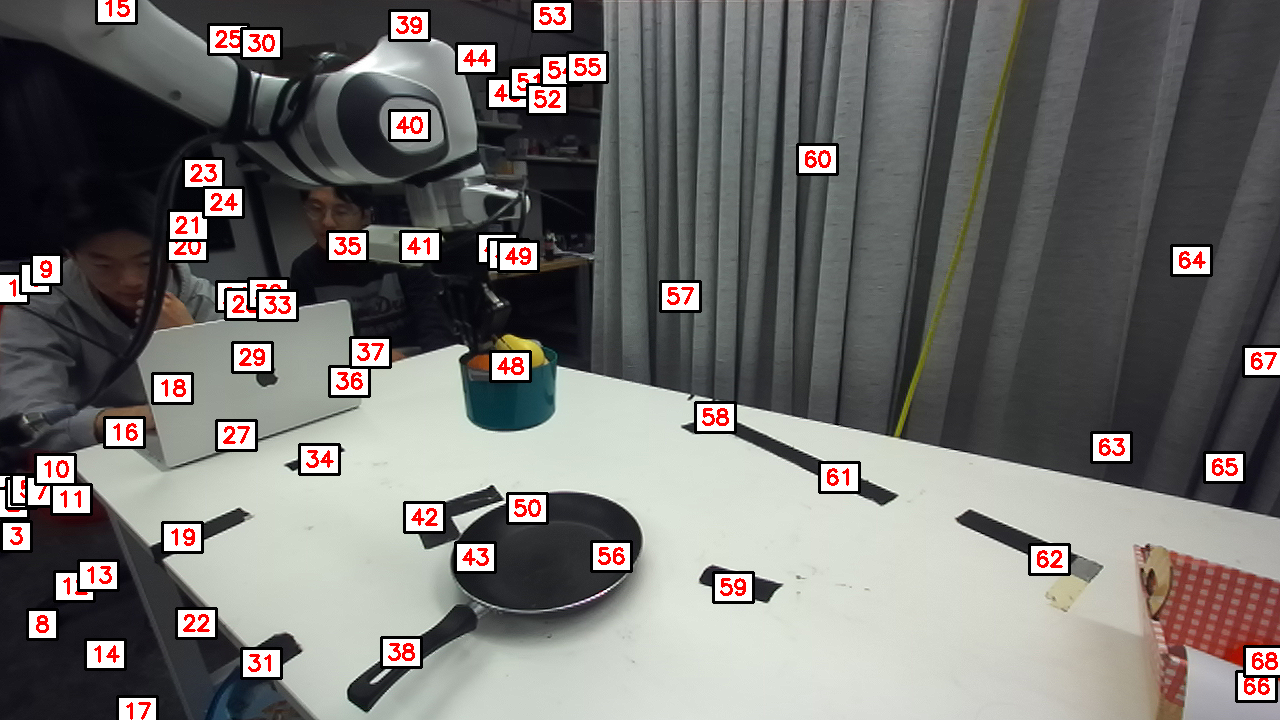}
\caption{\axadd{Keypoints proposed by ReKep for the \textit{Pour Food from Bowl into Pan} task.}}
\label{fig:pour_food_rekep_keypoints}
\end{figure}

\begin{lstlisting}[style=pythoncode, caption=\axadd{The constraints generated by ReKep for the \textit{Pour Food from Bowl into Pan} task ensure that the bowl is tilted at an angle of $45^{\circ}$ with respect to the $z$-axis to facilitate pouring (Line 18). However, this angle is insufficient to effectively pour the food out of the bowl.}, label=code:pour_food_rekep_constraints]
def path_constraint1(end_effector, keypoints):
    """
    Ensure the robot continues to hold the bowl during the pouring process.
    This can be achieved by keeping the end-effector aligned with the bowl's keypoint (e.g., keypoint 48).
    """
    cost = np.linalg.norm(end_effector - keypoints[48])
    return cost
    
def subgoal_constraint1(end_effector, keypoints):
    """
    Ensure the bowl is tilted to pour the object into the pot.
    This can be achieved by ensuring the vector formed by two keypoints on the bowl (e.g., keypoints 48 and 49) 
    is at a specific angle with respect to the z-axis.
    """
    bowl_vector = keypoints[49] - keypoints[48]
    z_axis = np.array([0, 0, 1])
    angle = np.arccos(np.dot(bowl_vector, z_axis) / (np.linalg.norm(bowl_vector) * np.linalg.norm(z_axis)))
    desired_angle = np.pi / 4  # Tilt the bowl by 45 degrees
    cost = np.abs(angle - desired_angle)
    cost = np.linalg.norm(bowl_vector)
    return cost
\end{lstlisting}

\paragraph{\textit{Pour Food from Bowl into Pan}}\axadd{Figure~\ref{fig:pour_food_rekep_keypoints} illustrates the keypoints proposed by ReKep, while Code Snippet~\ref{code:pour_food_rekep_constraints} presents the corresponding constraints. In the pouring task, while ReKep correctly establishes a rotation constraint, it underestimates the numerical value of the required rotation. As a result, the bowl is only slightly tilted at $45^{\circ}$, failing to achieve the intended pouring motion to empty the bowl. While ReKep demonstrates the Pour Tea task in its paper, the prompt used for the VLM in the publicly released implementation includes helpful guidance on constraint construction for this task, such as suggesting that ``the teapot must remain upright to avoid spilling''. This additional guidance may have enhanced ReKep's performance on the task. While pouring tea requires only a slight tilt, pouring food from a bowl into a pan demands a significantly larger tilt, something the VLM fails to reason about effectively.}

\end{document}